%% file: main.tex
\PassOptionsToPackage{dvipsnames}{xcolor}
\documentclass{adobe_research}
\usepackage[T1]{fontenc}
\usepackage{lmodern}
\usepackage{newtxtext}

\usepackage{amsmath,amssymb}
\renewcommand{\eqref}[1]{equation~\ref{#1}}
\usepackage{array}
\usepackage{adjustbox}
\usepackage{algorithm}
\usepackage[noend]{algpseudocode}
\usepackage{xspace}

\newcommand{\name}{AgiWorld\xspace}

\title{\name: An \underline{\smash{Ag}}ent\underline{i}c Video \underline{World} Model for Building Persistent Worlds}

\author[1]{Shaoheng Fang}
\author[2]{Yiqun Mei}
\author[2]{Hanwen Jiang}
\author[2]{Xin Sun}
\author[2]{Haitian Zheng}
\author[2]{Hao Tan}
\author[2]{Kalyan Sunkavalli}
\author[1]{Georgios Pavlakos}
\author[1]{Qixing Huang}
\author[2]{Yiwei Hu}
\affiliation[1]{\textbf{The University of Texas at Austin}}
\affiliation[2]{\textbf{Adobe Research}}

\abstract{\input{section/0_abstract}}

\date{November 2nd, 2026}
\adobedata[Project Page]{\url{https://bshfang.github.io/agiworld/}}

\begin{document}

\maketitle
\suppressfloats[t] % no figure above the title box on page 1

\input{section/1_introduction}

\input{section/3_framework_overview}

\input{section/6_method_agent}

\input{section/4_method_world_model}

\input{section/5_data_training}

\input{section/7_experiment}

\input{section/2_related_works}

\input{section/8_conclusion}

\bibliographystyle{assets/plainnat}
\bibliography{references}

\input{section/appendix}

\end{document}

%% file: section/0_abstract.tex
Recent interactive video world models generate visually compelling videos from a single image, following user-provided camera motion and text prompts. However, such low-level control demands substantial user effort to realize a high-level intent, and the generated world struggles to remain persistent as the exploration proceeds. In this work, we present \textbf{\name}, an agentic video world model that explores and builds a persistent world from a single image and a free-form instruction. For high-level planning, our agent \textbf{\name-Agent} designs the world and the journey through it in a world state and directs the exploration. For low-level realization, our video world model \textbf{\name-WM} generates the video chunk by chunk and stores it with its geometry in a spatial memory, from which the agent updates its world state. This closed loop keeps the world faithful to the user's intent and keeps the agent's world state consistent with the generated content throughout the exploration. We co-design the two levels around the shared spatial memory and an interface of metric camera trajectories and structured prompts. In \name-WM, to handle occlusion and accumulated geometric errors in the spatial memory over long rollouts, we propose occlusion-aware history selection, which retrieves past views by surface visibility, and reliability-based memory maintenance, which prefers reliable geometry and earlier observations. \name-WM achieves state-of-the-art overall performance on world revisiting and exploration benchmarks, and qualitative results and ablations show that \name builds persistent worlds from free-form instructions over minute-long journeys. Persistence thus arises not from the video model alone but from its closed loop with the agent.

%% file: section/1_introduction.tex
\section{Introduction}

\begin{figure}[t]
\begin{center}
\includegraphics[width=1.0\linewidth]{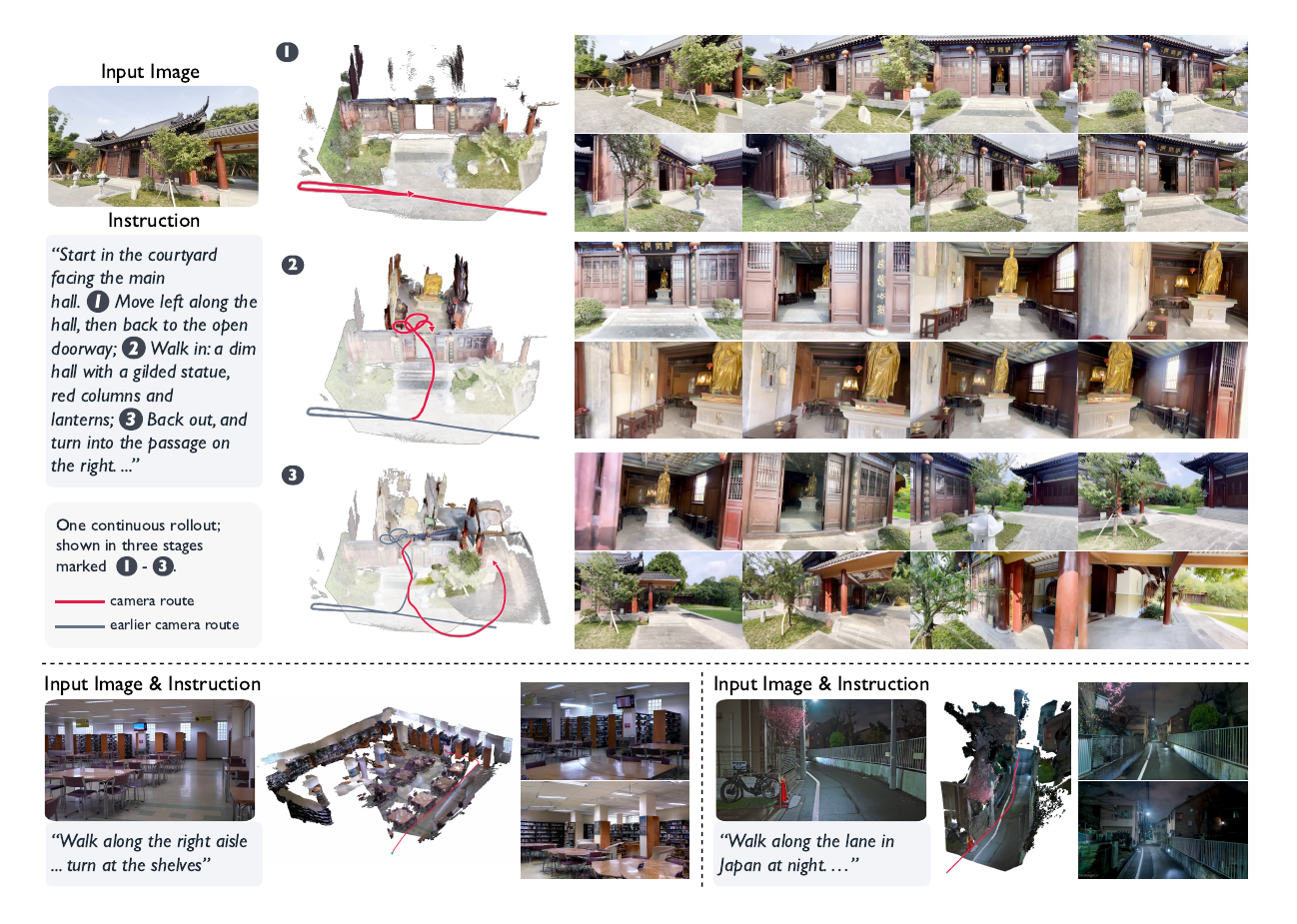}
\end{center}
\caption{\textbf{Persistent worlds built by \name.} Given an input image and a free-form instruction, \name-Agent designs the world and the journey through it, and directs \name-WM with a camera trajectory and a text prompt for each chunk. \name-WM generates the video autoregressively and uses a point cloud as its spatial memory, which in turn is fed back to \name-Agent for further exploration.
We show the rendered spatial memory with the camera route planned by the agent, and frames sampled from the long rollout videos. The results demonstrate the 3D persistence of the generated worlds as well as semantically meaningful exploration routes of the agent.}
\label{fig:teaser}
\end{figure}

A world is more than what is currently visible. It has places beyond the current view, spatial relationships between them, and content that should remain coherent as one moves through it. Building an explorable world therefore requires two capabilities: \emph{generating} what the camera sees and \emph{reasoning} about the world beyond the camera.

Recent video world models have made remarkable progress on the first capability~\citep{hong2025relic, team2026advancing, he2025matrixgame2}. Given an image, a camera trajectory, and a text prompt, they can generate realistic video as the camera moves through a scene. By generating successive video chunks, these models can support interactive exploration over increasingly long horizons. However, they do not decide what world should be explored. The camera trajectory and local content are still specified by the user, while content outside the current view is otherwise left to the model's generative prior. As a result, turning a high-level intent into a coherent journey still requires the user to continuously decide where to go and what should appear next.

We ask a different question: can a world model build and explore a world from a high-level instruction by itself? Doing so requires reasoning at a level above individual video chunks. Before entering an unseen region, the system must decide what should be there and how it connects to what has already been seen. After generating it, the system must treat the generated content as part of the world and reason from it thereafter. World building is therefore inherently a closed-loop process: \emph{plan the world, generate part of it, observe what was generated, and plan again}.

We introduce \textbf{\name}, an agentic video world model built around this loop. Given a single image and a free-form instruction, \name autonomously designs a world, plans a journey through it, and realizes the journey as a long video (Figure~\ref{fig:teaser}). It separates world building into two complementary components, an agent for high-level planning and a video world model for low-level realization. \textbf{\name-Agent} reasons about the world at a high level, maintaining a world state that contains both what has already been observed and what it has designed beyond the current view. \textbf{\name-WM} realizes this plan through camera-controlled video generation and maintains a spatial memory of what has actually been generated. After every generated chunk, the agent reads this memory, reconciles its plan with the generated world, and decides what should happen next.

For high-level planning, \textbf{\name-Agent}, built on a multimodal language model, keeps a world state that records the spatial layout of the world and the stages of the journey. The layout contains confirmed elements, which have already appeared in the video, and imagined elements, which the agent has designed but the camera has not yet reached. For each chunk, a director in \name-Agent turns the next stage of the journey into a camera move, such as walking through a door, turning to look around a room, or returning to a place seen before, and writes it as a metric camera trajectory and a structured text prompt. After each chunk, the agent confirms and corrects the imagined elements that were realized, retires those that the video contradicts, and extends the design ahead.

For low-level realization, we propose \textbf{\name-WM}, a novel camera-conditioned autoregressive video generation model that follows the director's trajectories and prompts accurately and keeps the generated world consistent over long rollouts. We condition the model on metric camera trajectories through a camera adapter and a relative camera positional encoding, PRoPE~\citep{li2025prope}.
For spatial memory, prior work compresses historical context~\citep{zhang2025framepack} or retrieves past views using camera overlap~\citep{yu2025contextasmemory} and spatial indexing~\citep{li2025vmem}. Reusing the geometry stored in our spatial memory raises two problems over long rollouts. A past view that overlaps the new camera may show surfaces that are occluded from it, and the estimated geometry accumulates errors. We therefore introduce occlusion-aware history selection, which retrieves past views by the surfaces visible from the new camera, and reliability-based memory maintenance, which prefers reliable geometry and earlier observations of each surface.

Our main contributions are as follows:
\begin{itemize}
\item We propose \name, a novel agentic video world model that couples an agent for high-level planning with a video world model for low-level realization, with the two levels co-designed.
\item We introduce \name-Agent, which designs the world and the journey through it, directs the video world model to generate them, and revises this design in a closed loop.
\item We design \name-WM, a camera-conditioned autoregressive video generation model supported by novel occlusion-aware history selection and reliability-based memory maintenance.
\item Experiments show that \name-WM achieves state-of-the-art overall performance on world revisiting and exploration benchmarks, that its two memory components both improve long-term consistency, and that \name builds persistent worlds from free-form instructions over minute-long journeys.
\end{itemize}

%% file: section/3_framework_overview.tex
\section{System Overview}

\begin{figure}[t]
\begin{center}
\includegraphics[width=1.0\linewidth]{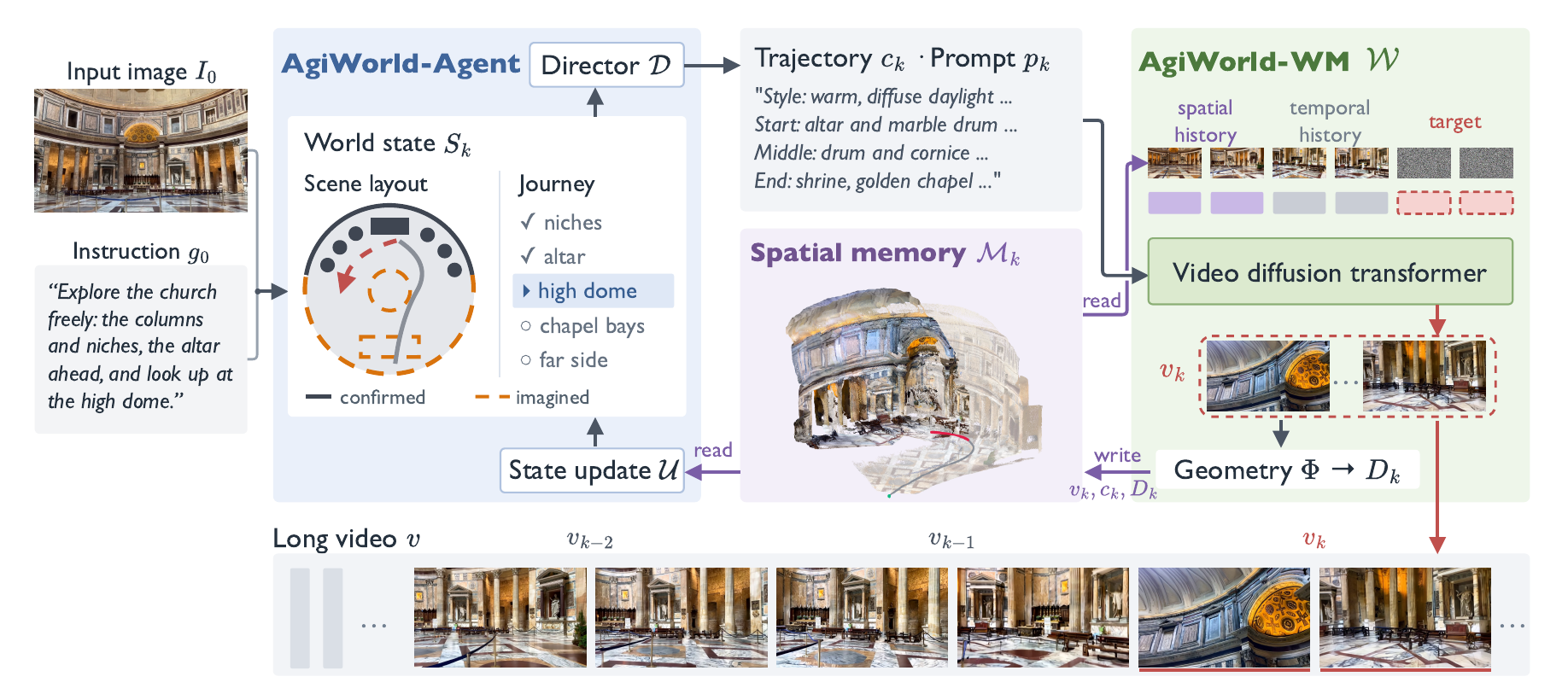}
\end{center}
\caption{\textbf{Overview of \name.} The director $\mathcal{D}$ turns the world state $S_k$ into $c_k$ and $p_k$. \name-WM generates $v_k$ from the spatial history read from its memory (purple, two of five shown) and the temporal history (gray), and writes $v_k$ with its depth $D_k$ into the memory, which the state update $\mathcal{U}$ reads to write $S_{k+1}$.}
\label{fig:framework_overview}
\end{figure}

\subsection{Problem Formulation}
\label{subsec:problem}

The input to \name is a single image $I_0$ and a free-form instruction $g_0$ written in natural language. The output is a long video $v=(v_1,\dots,v_K)$ that starts from the image, builds a world around it, and travels through this world as the instruction describes.
The video is generated chunk by chunk by a camera-controlled video world model $\mathcal{W}$. For chunk $k$, the model is conditioned on a camera trajectory $c_k=\{\mathbf{T}_{k,f}\}_{f=1}^{n}$, where $\mathbf{T}_{k,f} \in SE(3)$ is the camera pose of the $f$-th frame of the $n$ frames in chunk $k$, a text prompt $p_k$ that describes the content of the chunk, and the video generated so far:

$$
v_k=\mathcal{W}(v_{<k},\,c_k,\,p_k),
$$

where $v_{<1}=I_0$, and all camera poses are metric and expressed in the coordinates of the first frame $I_0$. Unlike $I_0$ and $g_0$, the trajectories and prompts are not given. They have to be constructed from $g_0$ as the journey proceeds, because where the camera can go and what it should see depend on what has been generated so far.

We call the generated world \textit{persistent} when two conditions hold. \textit{Visual persistence} concerns the appearance and geometry of the generated frames. Whenever the camera returns to a region that has been generated before, the new frames agree with the earlier ones in appearance and geometry, regardless of how many chunks have been generated in between, and the style of the scene does not drift over the journey.
\textit{Semantic persistence} concerns the content of the world and the journey. The world unfolds as the instruction $g_0$ describes, the scene stays coherent as it grows, and the journey through it is one continuous story rather than a series of unrelated views.

\subsection{Closed-Loop Generation}
\label{subsec:closed_loop}

We construct the trajectories and prompts with an agent, \name-Agent, which runs in a closed loop with the world model $\mathcal{W}$, \name-WM (Figure~\ref{fig:framework_overview}). The world model is responsible for visual persistence through its spatial memory, and the agent is responsible for semantic persistence through its world state. Together, the agent and the world model form \name, an agentic video world model.

The world model organizes its history $v_{<k}$ into a spatial memory. After a chunk is generated, a geometry estimator $\Phi$ estimates the depth $D_k=\Phi(v_k,\,c_k)$ for its frames. The memory $\mathcal{M}_k=\{(v_j,\,c_j,\,D_j)\}_{j<k}$ holds all chunks generated before chunk $k$, together with their trajectories and depths. The world model and its memory can then be written as

$$
\begin{aligned}
v_k &= \mathcal{W}(\mathcal{M}_k,\,c_k,\,p_k), \\
D_k &= \Phi(v_k,\,c_k), \\
\mathcal{M}_{k+1} &= \mathcal{M}_k\cup\{(v_k,\,c_k,\,D_k)\},
\end{aligned}
$$

starting from $\mathcal{M}_1=\{(I_0,\,\mathbf{T}_0,\,\Phi(I_0))\}$, where $\mathbf{T}_0$ is the camera pose of $I_0$.

The agent keeps the world in a high-level world state $S_k$ rather than pixels. The state records the layout of the world in two parts, what has been generated and what the agent has designed but not yet generated, together with the progress of the video journey. The agent has two main functions. A director $\mathcal{D}$ directs the world model: based on the world state and the instruction, it decides the trajectory and the prompt of the next chunk. A state update $\mathcal{U}$ reads the memory of the world model after the chunk has been generated, updates the world state, and designs (or redesigns) the world and the journey ahead:

$$
\begin{aligned}
(c_k,\,p_k) &= \mathcal{D}(S_k,\,g_0), \\
S_{k+1} &= \mathcal{U}(S_k,\,\mathcal{M}_{k+1},\,g_0),
\end{aligned}
$$

starting from $S_1=\mathcal{U}(\varnothing,\,\mathcal{M}_1,\,g_0)$, which reconstructs the visible scene from the input image into an initial world state and designs the journey and the world beyond it. Before the state update, the director reviews each generated chunk, and a chunk that fails the review is removed from the memory and generated again from a revised trajectory and prompt. The loop runs until the agent judges that the instruction $g_0$ has been carried out, so the number of chunks $K$ is not fixed in advance. In Section~\ref{sec:agentic_framework}, we detail the world state representation, its geometric grounding, the director with its review, and the state update. In Section~\ref{sec:method}, we detail the accurate camera following of the world model, occlusion-aware history selection, and reliability-based memory maintenance. In Section~\ref{sec:data_and_training}, we detail the data pipeline and the training stages of the world model.

%% file: section/6_method_agent.tex
\section{\name-Agent: Closed-Loop World-Building Agent}
\label{sec:agentic_framework}

The goal of \name-Agent is a journey through one world that unfolds as the instruction describes and stays coherent as it grows. This goal cannot be reached by choosing the trajectory and the prompt one chunk at a time. The world beyond the frame has to be decided before the camera reaches it, the places that the journey visits have to fit into one spatial layout, and every chunk has to advance the same journey. Since the world takes shape only as it is generated, these decisions also have to be revised after every chunk. \name-Agent therefore keeps a persistent world state that holds the layout of the world and the stages of the journey (Section~\ref{subsec:agent_state}), and works with this state through geometric tools that do all geometric computation for it (Section~\ref{subsec:agent_geometry}). A director derives the trajectory and the prompt of each chunk from this state and reviews the generated chunk (Section~\ref{subsec:agent_director}), and a state update revises the plan from what was generated and extends it ahead of the camera (Section~\ref{subsec:agent_update}). Figure~\ref{fig:agentic_detailed_framework} in Appendix~\ref{appendix:agent_framework} shows one iteration of this loop, and the appendix gives the implementation details.

\begin{figure}[t]
\begin{center}
\includegraphics[width=1.0\linewidth]{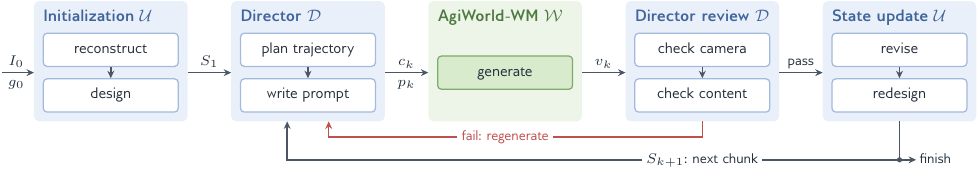}
\end{center}
\caption{\textbf{Pipeline of \name-Agent.} The state update $\mathcal{U}$ builds the initial world state $S_1$ from $I_0$ and $g_0$. For each chunk, the director $\mathcal{D}$ plans $c_k$ and $p_k$, \name-WM generates $v_k$, and the director reviews it. A chunk that fails is generated again, and a chunk that passes is used by $\mathcal{U}$ to write $S_{k+1}$.}
\label{fig:agent_framework_overview}
\end{figure}

\subsection{High-Level World State}
\label{subsec:agent_state}

The state $S_k$ is a compact, high-level description of the world, separate from the world model's memory $\mathcal{M}_k$ and expressed in the coordinate frame of $I_0$. It consists of two parts. The first is a scene layout: a set of named elements, including floors, walls, openings, and objects, each with a coarse footprint, a height, and a short appearance, and labeled either \emph{confirmed}, meaning generated and observed, or \emph{imagined}, meaning designed by the agent but not yet reached by the camera. The second describes the journey: a fixed visual style for the whole video, an ordered sequence of stages, each stating what the camera should show and the layout elements it involves, and the progress made through them so far. The state is coarse and semantic, which keeps it easy to confirm and extend as the world is generated, and its imagined elements let the agent design the world beyond the frame before the camera arrives. 

\subsection{Geometric Grounding}
\label{subsec:agent_geometry}

The world state layout shares its metric coordinate frame with the camera poses and the memory of the world model, so a trajectory can be planned on the layout and every element can be checked against the generated geometry. The agent, however, is built on a multimodal language model, which can reason about a scene but cannot compute with point clouds or camera poses by itself. All geometric computation is therefore done by a set of deterministic tools that every part of the agent shares. Before each query to the language model, the tools prepare what it reads: images drawn in the frame of the layout with the same conventions, such as a top-down map of the generated point cloud on a metric grid with the layout and the camera path over it, or the layout projected into the frames of a chunk, together with quantities such as the visibility of each element along a planned trajectory or the height of the camera above the generated ground. After each query, the same tools check what the model wrote, from layout elements to waypoints, and return it with its errors when it fails.

\subsection{Director}
\label{subsec:agent_director}

The director turns the world state into the trajectory and the prompt of each chunk, $(c_k, p_k) = \mathcal{D}(S_k, g_0)$. A few waypoints are placed on the layout toward what the active stages should show, following camera practices that suit what the world model can generate, and solved into the per-frame metric poses $c_k$ in the coordinate frame of $I_0$, matching the camera conditioning of Section~\ref{subsec:method_camera}. The prompt is then derived from this trajectory. The layout is projected along the planned poses to find the elements in view at the start of the chunk, the elements that enter or grow during it, the elements that leave the view, and the elements in view at its end. The director writes these elements into the structured format of the training captions (Section~\ref{sec:data_and_training}), a style line fixed for the whole journey followed by the start, the middle, and the end of the chunk. The camera motion and the prompt therefore describe the same content, and consecutive chunks follow one plan. Rather than committing the whole journey at once, the director plans a bounded window ahead and extends it as the rollout proceeds.

After each chunk, the director reviews the generated chunk against the plan. The review checks whether the elements named in the prompt and those that the next stages rely on were generated as planned, and measures the camera on the geometry that the world model estimated, such as its height above the generated ground and the nearest solid ahead. A chunk whose camera fails these measurements, or that shows something else in place of an element named in the instruction, is removed from the memory and generated again from a redesigned trajectory and prompt.

\subsection{World Design and State Update}
\label{subsec:agent_update}

The state update $S_{k+1} = \mathcal{U}(S_k, \mathcal{M}_{k+1}, g_0)$ both designs the world ahead of the camera and reconciles the state with what is generated, closing the loop between the agent and the world model. The agent designs the imagined layout and the journey for the stretch to come, so the plan always extends beyond what has been generated. Stages are closed once their content has been shown, and new stages are appended when the remaining plan runs short. The imagined layout is also kept walkable from the confirmed one, with imagined ground connected to confirmed ground and no imagined element crossing a confirmed wall, so the camera can always be routed into what the agent has designed.

Once a chunk is generated, the agent observes it directly from the world model's memory, reading the geometry the model has already estimated for it, so no separate perception step is needed. The update follows one principle: the generated video overrides the agent's design but not the user's instruction. The agent therefore updates the state, confirming the imagined elements that were realized, correcting their geometry, and retiring those the new content contradicts, so the layout agrees with the generated world. Retired elements are kept as a record so that they are not proposed again. A stage whose space the video contradicted is skipped, and the design continues around what was generated. After a major mismatch, the confirmed layout is rebuilt from the generated geometry. Elements seen in the input image or named in the instruction are the exception. They are not rewritten to match a generated chunk, and when the video shows something else in their place, the agent treats the generation as the error and keeps its prompts and route faithful to the instruction. The director then derives the trajectory and the prompt of the next chunk from the updated state.

%% file: section/4_method_world_model.tex
\section{\name-WM: Persistent Video World Model}
\label{sec:method}

The goal of \name-WM is to follow the trajectory and the prompt of every chunk accurately and to keep the generated world consistent over the whole rollout, so that a region looks the same whenever the camera returns to it. Existing camera-controlled world models generate long videos autoregressively, conditioning each iteration on recent frames~\citep{li2025gamecraft}, a compressed history~\citep{hong2025relic}, or past views retrieved by camera overlap~\citep{wang2026matrixgame3}. Our model follows the same autoregressive scheme, in chunks of five seconds that match the unit at which the agent acts, and keeps the generated frames with their metric geometry as memory (see Section~\ref{subsec:method_ar}).
Accurate camera following is the foundation of reliable memory. We condition the model on metric cameras through a camera adapter and a relative camera positional encoding (see Section~\ref{subsec:method_camera}).
The difficulty then lies in keeping the memory consistent over a long rollout, despite occlusion and error accumulation in the estimated geometry. To solve these problems, we propose occlusion-aware history selection, which retrieves and injects only the history visible from the new camera, and reliability-based memory maintenance, which prefers reliable geometry and gives earlier observations precedence (see Section~\ref{subsec:method_memory}).

\subsection{Autoregressive Video Generation}
\label{subsec:method_ar}

We generate the video $v=(v_1,\dots,v_K)$ chunk by chunk with a bidirectional video diffusion model. For chunk $k$, the history is placed in front of the target latents of $v_k$ as clean context, and the target is denoised conditioned on this history, on the trajectory $c_k$, and on the prompt $p_k$. The history consists of two parts, a spatial history and a temporal history.
The spatial history consists of five frames retrieved from the memory $\mathcal{M}_k=\{(v_j,\,c_j,\,D_j)\}_{j<k}$. Since all poses and depths share the coordinate frame of $I_0$, the memory forms a point cloud of the generated world, and we retrieve the frames whose points cover what the trajectory $c_k$ is about to see.
The temporal history consists of the last frames of $v_{k-1}$, encoded jointly with the chunk so that the video continues without a seam. In our design, the temporal history is short and is responsible only for the continuity of motion and appearance between consecutive chunks. Long-term persistence relies on the spatial history, which is retrieved from the entire memory by geometry rather than by recency. The history context has a fixed token budget independent of the rollout length, which allows generation to continue indefinitely. After $v_k$ is generated, its frames are decoded, the geometry estimator gives $D_k=\Phi(v_k,\,c_k)$, and the memory grows to $\mathcal{M}_{k+1}=\mathcal{M}_k\cup\{(v_k,\,c_k,\,D_k)\}$. For the first chunk, the memory holds only the input image, and the chunk is generated from $I_0$ alone.

\subsection{Accurate Camera Following}
\label{subsec:method_camera}

The world model has to follow the trajectory $c_k$ accurately. The poses in $c_k$ are metric and in the coordinates of $I_0$, without the per-clip normalization common in camera-conditioned video models that discards the absolute scale of the motion. The trajectory enters the model through two paths. A camera adapter represents the relative pose between consecutive frames as a per-pixel ray embedding and modulates the features of the generated tokens in every block with a learned scale and shift. It acts on the generated tokens only, giving each frame a signal of how the camera has moved since the previous frame. However, a generated token also needs to know the relative camera to every history token it attends to, including memory frames from distant viewpoints.
We therefore add PRoPE~\citep{li2025prope} to the self-attention. Let $\mathbf{P}_i = \mathbf{K}\mathbf{T}_i^{-1}$ be the projection of the camera of token $i$, with $\mathbf{T}_i$ its pose and $\mathbf{K}$ the intrinsics. In a second attention branch in every block, queries are transformed by $\mathbf{P}_i^{\top}$ and keys and values by $\mathbf{P}_j^{-1}$, so the attention logit depends only on the relative projection $\mathbf{P}_i \mathbf{P}_j^{-1}$ between two tokens $i$ and $j$, not on the choice of world coordinates. This encoding covers the full sequence and is the only camera signal that history frames receive.

\subsection{Spatial Memory for Persistent Generation}
\label{subsec:method_memory}

The spatial memory is what keeps the world persistent across a long rollout. Every frame in $\mathcal{M}_k$ is back-projected from its depth into metric world points, so the memory is a point cloud of the generated world in which every point is linked to the frame that observed it.
Making this memory reliable requires solving three problems: (i) how to retrieve frames from the memory without letting occluded surfaces contaminate the spatial history, (ii) how to inject the geometric information carried by the memory into the generation of new views, and (iii) how to keep the memory trustworthy at inference, where it is built from the model's own outputs and depth errors can compound across chunks. The first two are applied in both training and inference, and the third arises only at inference.

\textbf{Occlusion-aware history selection.}
From the point cloud of the entire generated world, we retrieve only a fixed small set of frames as the spatial history, the ones most relevant to the trajectory $c_k$. Previous works may select them by camera overlap~\citep{wang2026matrixgame3}, but overlap compares only camera poses and can keep frames whose surfaces are hidden from the target views. A more geometric method is to render the point cloud into the target views with a shared z-buffer~\citep{yu2024viewcrafter} and keep the frames whose points survive. But a point cloud is points, not surfaces. It has no surface orientation, so a point on the back of an object counts the same as one on its front. A point cloud also leaves gaps between its points, so a nearer surface cannot fully hide what lies behind it. When the camera turns from the front of an object to its back, the frames worth keeping change entirely, yet a point-based test still treats the now-hidden front as visible. We fix both by giving each point a normal and rendering the points as small surfels. A memory point then counts as visible from the target views along $c_k$ only when its normal faces the camera and no nearer surface hides it. Concretely, the normal $\mathbf{n}_{\mathbf{x}}$ of a memory point $\mathbf{x}$ is estimated from its depth map. In a target view $r$ with camera center $\mathbf{o}_r$, the front-facing points are rasterized as surfels, the quads spanned by four neighboring pixels of the depth map, which fills the gaps between points and gives at each pixel the depth $d_r$ of the nearest front-facing surface. A point is visible when it faces the camera and its own depth $z_r(\mathbf{x})$ reaches this surface rather than lying behind it,
\begin{equation}
\mathrm{vis}_r(\mathbf{x})=\big[\,\mathbf{n}_{\mathbf{x}}\cdot(\mathbf{o}_r-\mathbf{x})>0\,\big]\,\big[\,z_r(\mathbf{x})\le(1+\delta)\,d_r(\pi_r(\mathbf{x}))\,\big],
\label{eq:vis}
\end{equation}
where $[\cdot]$ equals $1$ when the enclosed condition holds and $0$ otherwise, $\pi_r$ is the projection into view $r$, and the tolerance $\delta$ keeps a surface from occluding itself. A point counts as visible along $c_k$ when it is visible from any of six target views spaced evenly along the trajectory and ending at its last frame. Given this test, we build the spatial history by a greedy set cover over the target pixels of $c_k$. The last frame of $v_{k-1}$ is already in the sequence as part of the temporal history, so we start from the target pixels its visible points cover, and repeatedly add the memory frame whose visible points cover the most target pixels not yet covered, until five frames are selected. The selected frames therefore complement one another instead of repeating one viewpoint.

\textbf{Geometry correspondence injection.}
The spatial history selected is placed in the sequence, but this alone leaves the token-level correspondence between the memory latents and the target latents implicit. Conditioning on a frame warped into the target view~\citep{seo2024genwarp} or on a rendered point cloud~\citep{huang2025voyager} would make it explicit, but the model then reproduces the depth errors, holes, and stretched regions of these images in the generated frames. We instead inject an index rather than the source appearance, following coordinate-based geometry conditioning~\citep{bai2026pefield4d, shen2026lyra2}.
Concretely, we use two kinds of sources: the spatial history frames and the last frame of the temporal history. From each source, we warp a per-pixel index map and depth into every generated frame, linking each generated pixel that a source reaches to its source pixel. The index map stores at every pixel its normalized image coordinates and a code that identifies the source, and the warp also records the depth of each warped point in the target camera. Source pixels at depth discontinuities are not warped, and target pixels that no source reaches are left empty. We use the same occlusion-aware visibility test for retrieval and injection, so the two stay consistent. In each transformer block, an encoder turns this correspondence into a bias on the query and key inputs of the generated tokens, steering each token toward the memory it matches.

\textbf{Reliability-based memory maintenance.}
Both mechanisms above rely on a memory whose geometry is accurate and consistent. In training this is guaranteed: the geometry of each training video is estimated offline over its full length and filtered for quality, so the memory is stable. At inference, however, the memory is built incrementally from the model's own outputs. A small drift in camera following leaves a chunk's geometry slightly inconsistent, the next chunk is generated on top of it and drifts further, and the error compounds, accumulating over the rollout.
We therefore extend the occlusion-aware visibility test into a memory-maintenance step at inference to decide which pixels the model should trust across the many history frames and the point cloud they form. We introduce two judgments for this, on a pixel's reliability and on its generation time.
The reliability judgment tests three conditions for each point: (i) its confidence from the geometry estimator is above a percentile of the confidence over the memory, (ii) its incidence angle is below $\theta_{\max}$, so it was observed close to head-on, and (iii) its distance from the observing camera is below $z_{\max}$. Reliable points are used first in both retrieval and injection, and less reliable ones fill only the gaps they leave. An unreliable point is deprioritized but never discarded, so it still contributes to the occluding surface $d_r$, and an untrusted surface still hides what lies behind it.

Moreover, the generation-time judgment trusts earlier content. In a generated world there is no ground truth to check a chunk against, and consistency can only mean agreement with what the model has already shown. Earlier content is therefore the reference by definition, and a later point that contradicts it is the error, whatever its own confidence. A later point $\mathbf{x}$ conflicts with an earlier view $u$ when it lies clearly in front of the surface that view observed, $z_u(\mathbf{x})<(1-\delta)\,d_u(\pi_u(\mathbf{x}))$, that is, where the view saw empty space. When the earlier view observed that location reliably, we keep the earlier observation and drop the later, drifted point from retrieval, from the correspondence, and from the occluding surface. Later generation can then extend the world but not overwrite it, which curbs drift. In our implementation, $\theta_{\max}=70^\circ$, $z_{\max}=30$ m, and $\delta=0.05$. Together, these mechanisms let the model treat its own generated history as a reliable memory rather than an accumulating source of error, so the world stays consistent over long rollouts.

%% file: section/5_data_training.tex
\section{Training Data and Training Strategy}
\label{sec:data_and_training}

\begin{figure}[t]
\begin{center}
\includegraphics[width=0.98\linewidth]{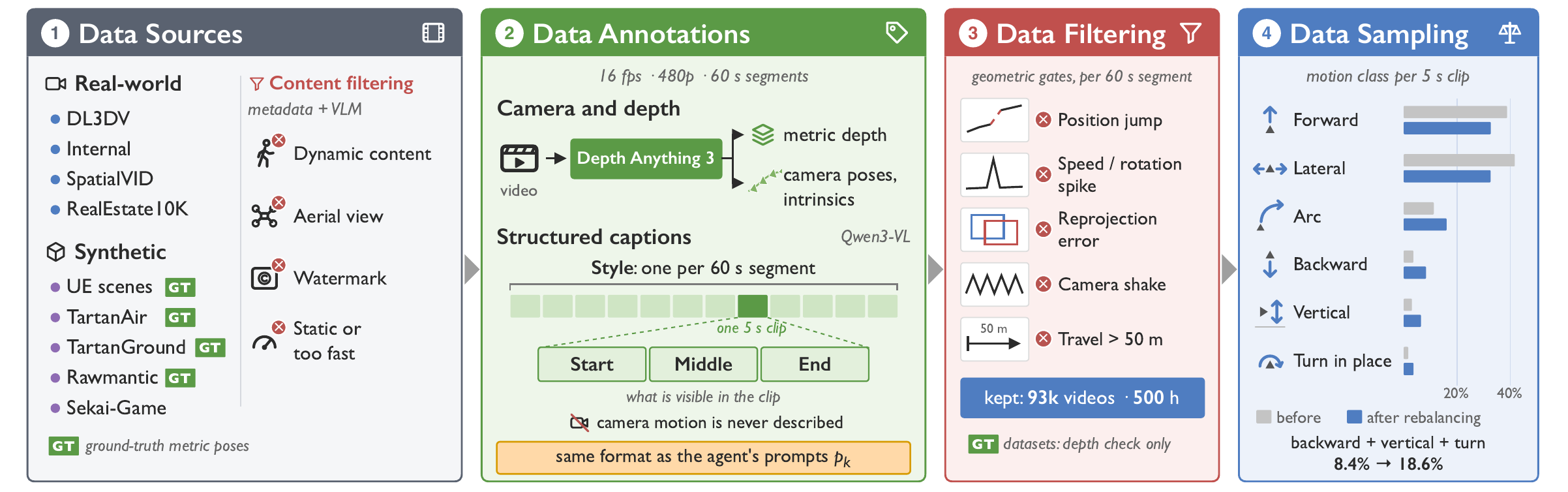}
\end{center}
\caption{Data pipeline of \name-WM. (1)~Data sources: four real-world and five synthetic datasets, filtered with metadata and a VLM. GT marks the datasets with ground-truth metric poses. (2)~Data annotations: metric depth, camera poses and intrinsics from Depth Anything 3~\citep{lin2025depth}, and a structured caption for every 5-second clip from Qwen3-VL~\citep{bai2025qwen3}. (3)~Data filtering: geometric gates discard segments with unreliable pose and depth annotations. (4)~Data sampling: clips are labeled with a camera motion class, and rare classes are sampled more often. The bars show the share of each class before and after rebalancing.}
\label{fig:data_pipeline}
\end{figure}

\subsection{Training Data}
\label{subsec:training_data}

\name-WM needs to follow metric camera trajectories accurately, keep a memory of everything it has generated, and faithfully generate the scene that the structured prompt describes. To support this, the training data needs metric camera poses with diverse camera motion, long videos with per-frame depth that return to earlier viewpoints, and scene diversity.
We build our data pipeline (Figure~\ref{fig:data_pipeline}) for training data that meet these requirements.

\textbf{Data sources.} We combine nine datasets. 
DL3DV~\citep{ling2024dl3dv} and an internal collection of indoor and outdoor scenes provide long-term real-world captures in which the camera moves around or explores a scene. SpatialVID~\citep{wang2026spatialvid} and RealEstate10K~\citep{zhou2018stereo} provide real-world walking, driving, and indoor walkthrough videos that mostly move forward and contribute diverse in-the-wild scenes.
The remaining five datasets are synthetic. 
Videos rendered from Unreal Engine scenes have the most balanced camera motion, including turns and vertical movement. TartanAir~\citep{tartanair2020iros} and TartanGround~\citep{patel2025tartanground} provide fast drone flights and ground traversal through large simulated environments. Rawmantic~\citep{rawmantic2026data} provides rendered interiors, and Sekai-Game~\citep{li2026sekai} provides walking videos recorded in a game engine. All synthetic datasets except Sekai-Game come with ground-truth metric camera poses.

\textbf{Data annotations.}
All videos are resampled to 16 fps at 480p resolution and split into segments of at most 60 seconds, and every segment receives camera and depth annotations and captions.
Depth Anything 3~\citep{lin2025depth} is used to jointly estimate camera poses, intrinsics, and metric depth for each segment. For the synthetic datasets with ground-truth poses, we keep the poses and run the same model conditioned on them.
Every 5-second clip is captioned by Qwen3-VL~\citep{bai2025qwen3} in a structured format: a style description shared by the whole segment, followed by descriptions of what is visible at the start, in the middle and at the end of the clip. The captions describe the scene but never the camera motion, which enters the model only through the camera condition. The structure makes the instruction to the model explicit, so the generated video follows the caption more faithfully. It also gives the director a fixed format to write its prompts in (Section~\ref{subsec:agent_director}), which makes generation easier to control.

\textbf{Data filtering.} 
We filter the data in two stages for high-quality source videos and reliable geometry annotations. We first remove footage that is unsuitable for training: clips with moving people, vehicles or flowing water, aerial views, near-static or overly fast camera motion, and large watermarks. These are detected from dataset metadata and by a local VLM~\citep{bai2025qwen3}. 

After annotating the camera parameters and depth, we apply geometric gates to every segment and discard those with unreliable geometry annotations, since a single erroneous pose corrupts the coordinate frame of the entire segment. The gates evaluate the estimated trajectory and depth for (i) sudden jumps of the camera position, (ii) sudden changes in speed or rotation, (iii) low depth reprojection consistency between nearby frames, and (iv) shaking of the camera. Their thresholds are calibrated separately for each dataset, and datasets with ground-truth poses pass only a depth quality check. We also discard training samples whose camera moves more than 50 meters, because the relative camera transforms in PRoPE~\citep{li2025prope} grow with translation and such samples produce exploding gradients. After filtering, about 93k source videos and 500 hours of video remain.

\textbf{Data sampling.}
Accurate camera control requires following arbitrary camera trajectories, in any direction and with any motion pattern, so the training data has to cover camera motion of all kinds. In the collected videos, however, most clips move straight forward or sideways, and turning, backward, and vertical motion are rare. We therefore label every 5-second clip with a camera motion class computed from its trajectory, namely forward, backward, lateral, vertical, turning in place, or moving along an arc. During training, we sample clips of the rare motion classes more often, so that the different camera motions are more evenly represented.

\begin{figure}[t]
\begin{center}
\includegraphics[width=0.95\linewidth]{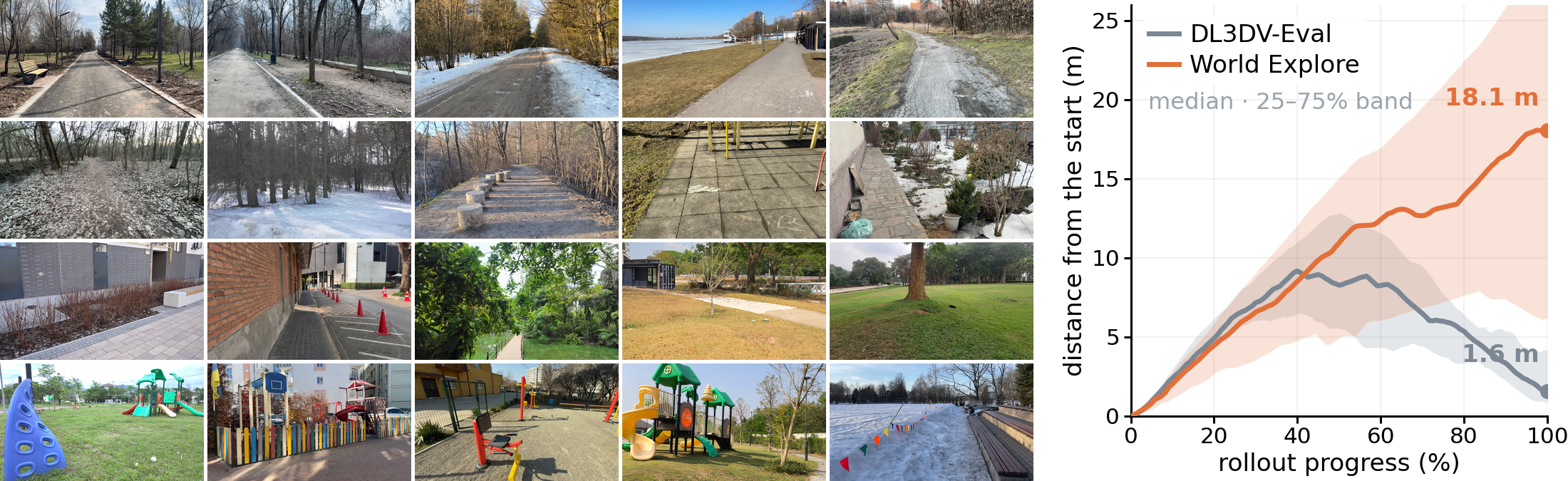}
\end{center}
\caption{World Explore benchmark. Left: the first frames of 20 of the 40 scenes. Right: distance of the camera from its starting position along the target trajectories of DL3DV-Evaluation and World Explore. DL3DV-Evaluation trajectories move away and come back to a median distance of 1.6\,m from the start, whereas World Explore trajectories keep moving away and end at a median distance of 18.1\,m.}
\label{fig:world_explore_benchmark}
\end{figure}
\subsection{Staged Training}

We build our model on Wan2.2-I2V-A14B~\citep{wan2025}, initialized from the LingBot-World checkpoint~\citep{team2026advancing}. The model is trained in three stages, each building on the previous.

\textbf{Stage 1: Metric camera adapter.} We fine-tune only the camera adapter, freezing the rest of the backbone, so the model follows metric camera trajectories. Camera-conditioned models are usually trained with per-clip-normalized poses, which discard the absolute scale of the motion. We instead condition on the metric relative poses between consecutive frames and train the adapter to interpret their absolute scale. We train bidirectionally on single-chunk clips with the base model's flow-matching objective, isolating accurate camera following before any memory is introduced.

\textbf{Stage 2: Memory-conditioned autoregressive training.} The second stage trains the model to generate autoregressively. As at inference, the spatial and temporal history is placed in front of the target chunk as clean context, and the loss is applied only to the generated frames. We add the relative camera positional encoding and the correspondence-injection encoder, both zero-initialized, and train them together with the self-attention, while the camera adapter from Stage 1 and the rest of the backbone stay frozen. The occlusion-aware history selection and the warped correspondences are parameter-free geometry, computed from the annotated geometry at training and from the estimated geometry of the generated frames at inference, so only the modules that inject them into attention are learned. Since the history at inference comes from the model's own drifting output, we add random noise and saturation to the clean context during training to keep the model robust to that drift.

\textbf{Stage 3: Few-step distillation.} Generating a chunk with the full-step model requires tens of denoising steps, which becomes expensive over the long rollouts of agent-driven world building. Following Phased DMD~\citep{fan2026phased}, we distill the high-noise and low-noise experts of the model to two steps each. We also distill classifier-free guidance, so each step is a single forward pass rather than two. A chunk is then generated in four forward passes.

%% file: section/7_experiment.tex
\section{Experiments}

\subsection{World Model Evaluation}

% Table 1: FID/VBench from the wm-eval metrics page 20260909_metrics_en (Ours on DL3DV from the 2026-09-15 Group D run); DL3DV Trans./Rot./Repr. from the Group D table (2 fps grid, 120 views, occupancy write band 0.30). Generator: paper_materials/03_experiments/table1_main/.

% One float holding two figures: with one top float per page, this is the only way to get
% both at the top of the same page. Two \caption commands give two figure numbers.
\begin{figure}[t]
\begin{center}
\includegraphics[width=1.0\linewidth]{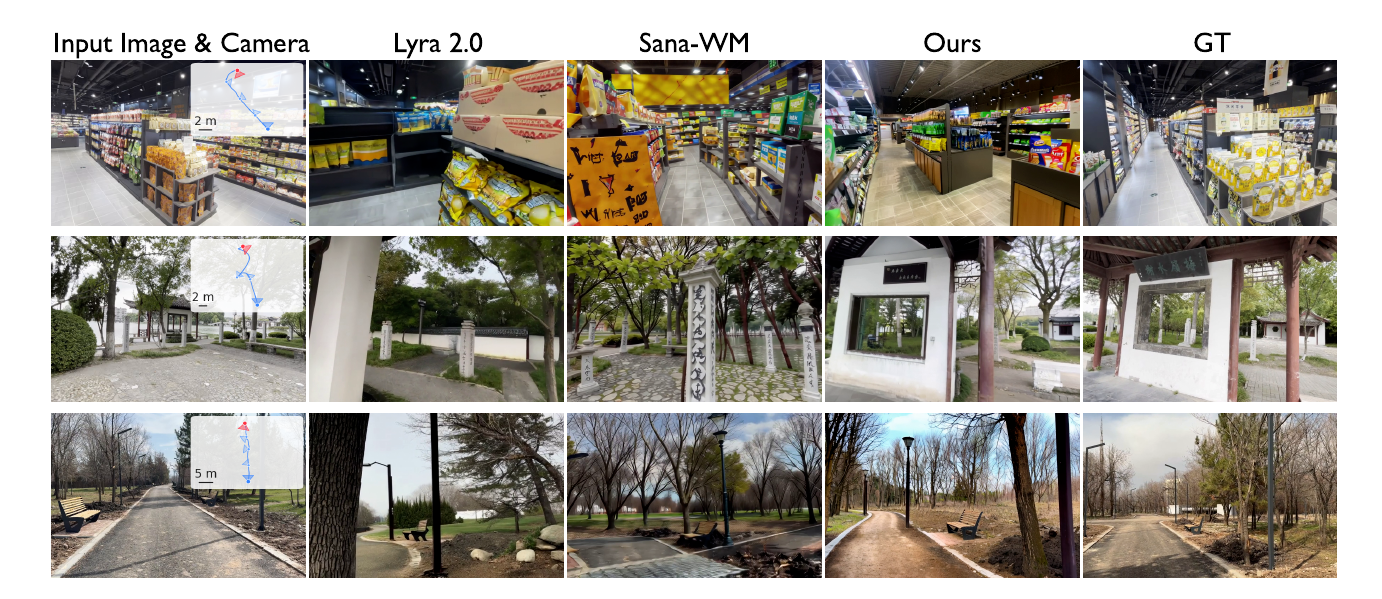}
\end{center}
\vspace{-2mm}
\caption{Camera-controlled long video generation comparison on DL3DV-Evaluation and World Explore. The first column shows the input image with a top view of the target camera trajectory.}
\label{fig:comparison_video_gen}
\vspace{10pt}
\begin{center}
\includegraphics[width=0.95\linewidth]{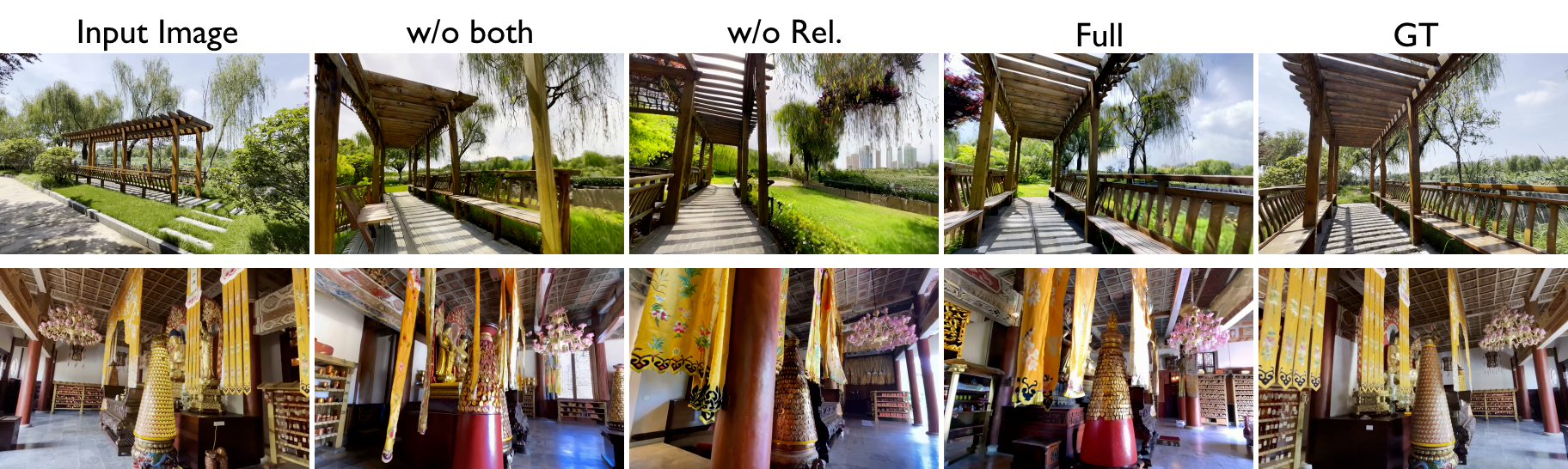}
\end{center}
\caption{Qualitative ablation of the two memory components. Each row shows one scene: the input image, the frames generated at one later time step by the variant without both components (w/o both), the variant without reliability-based memory maintenance (w/o Rel.) and the full model (Full), and the ground-truth frame at the same camera pose (GT).}
\label{fig:comparison_ablation}
\end{figure}

\begin{table}[t]
\caption{Quantitative comparison with camera-controlled world models on DL3DV-Evaluation (world revisiting) and World Explore (world exploration). Imag., Aes. and Text are the VBench imaging quality, aesthetic quality and text consistency scores. Trans., Rot. and Repr. are translation (m), rotation ($^\circ$) and depth reprojection errors. Best in \textbf{bold}, second best \underline{underlined}.}
\label{tab:main}
\centering
\scriptsize
\setlength{\tabcolsep}{1.3pt}
\begin{adjustbox}{max width=\textwidth}
\newcolumntype{C}{w{c}{19pt}}  % every metric column has the same width
\begin{tabular}{@{}l *{14}{C}@{}}
\toprule
 & \multicolumn{7}{c}{DL3DV-Evaluation} & \multicolumn{7}{c}{World Explore} \\
\cmidrule(lr){2-8} \cmidrule(lr){9-15}
Method & FID$\downarrow$ & Imag.$\uparrow$ & Aes.$\uparrow$ & Text$\uparrow$ & Trans.$\downarrow$ & Rot.$\downarrow$ & Repr.$\downarrow$ & FID$\downarrow$ & Imag.$\uparrow$ & Aes.$\uparrow$ & Text$\uparrow$ & Trans.$\downarrow$ & Rot.$\downarrow$ & Repr.$\downarrow$ \\
\midrule
SPMem~\citep{wu2025spatialmemory} & 90.26 & 49.72 & 0.420 & 0.186 & 2.98 & 6.39 & 0.085 & 137.78 & 49.96 & 0.397 & 0.163 & 3.91 & 4.00 & 0.072 \\
GEN3C~\citep{ren2025gen3c} & 41.11 & 60.07 & 0.498 & 0.213 & 1.07 & 1.94 & 0.031 & 63.39 & 53.84 & 0.435 & 0.187 & 2.05 & 3.64 & 0.052 \\
HY-WorldPlay~\citep{sun2025worldplay} & 81.34 & 56.12 & 0.395 & 0.152 & 3.08 & 5.36 & 0.049 & 103.03 & 54.50 & 0.378 & 0.127 & 4.01 & 6.04 & 0.095 \\
LingBot-World~\citep{team2026advancing} & 69.05 & 61.90 & 0.459 & 0.177 & 2.74 & 7.65 & 0.073 & 57.95 & 67.51 & \underline{0.471} & 0.181 & 3.36 & 4.35 & 0.075 \\
Matrix-Game 3.0~\citep{wang2026matrixgame3} & 46.50 & 70.09 & 0.477 & 0.191 & 2.09 & 6.64 & 0.073 & 51.73 & 69.35 & 0.462 & 0.177 & 2.46 & 2.02 & 0.065 \\
Lyra 2.0~\citep{shen2026lyra2} & \underline{25.54} & \underline{73.02} & 0.486 & \underline{0.232} & \underline{0.34} & \textbf{0.63} & \underline{0.021} & \underline{42.73} & \underline{72.05} & 0.461 & \underline{0.216} & \underline{1.32} & \underline{1.42} & \underline{0.037} \\
SANA-WM~\citep{zhu2026sana} & 48.32 & 72.01 & \textbf{0.547} & 0.215 & 1.28 & 2.69 & 0.038 & 65.31 & 70.15 & 0.467 & 0.195 & 1.87 & 1.61 & 0.045 \\
\midrule
\textbf{Ours} & \textbf{21.74} & \textbf{73.54} & \underline{0.524} & \textbf{0.236} & \textbf{0.29} & \underline{0.99} & \textbf{0.020} & \textbf{35.37} & \textbf{72.65} & \textbf{0.477} & \textbf{0.223} & \textbf{1.05} & \textbf{1.27} & \textbf{0.032} \\
\bottomrule
\end{tabular}
\end{adjustbox}
\end{table}

\textbf{Benchmarks.}
We evaluate long-term camera-conditioned video generation on two benchmarks. We evaluate \textit{world revisiting} on DL3DV-Evaluation~\citep{ling2024dl3dv}, whose 60-second camera trajectories circle the scene and return to where they started. Revisited regions therefore have to match what the model generated before, which tests its long-term memory. Besides, we introduce another benchmark, World Explore, to evaluate the model's capability of \textit{world exploration}, where the camera keeps moving into unseen parts of the scene, so most of the video is generated beyond the input image and has to stay consistent with the frames before it. World Explore consists of 40 real-world outdoor scenes held out from an internal collection and never used in training, each with a 60-second camera trajectory.
Section~\ref{subsec:world_explore_benchmark} describes the World Explore benchmark in detail.

\textbf{Metrics.}
We evaluate visual quality and text following by FID~\citep{heusel2017gans} and by the imaging quality, aesthetic quality, and text consistency scores of VBench~\citep{huang2024vbench}. We evaluate camera following and geometric consistency by estimating camera poses and depth from the generated video with Depth Anything 3~\citep{lin2025depth}, and report the mean translation and rotation errors of the estimated camera trajectory against the target trajectory (Trans., Rot.) and the depth reprojection error between frames one second apart (Repr.).

\textbf{Baselines.}
We compare with recent state-of-the-art camera-controlled long video generation models: SPMem~\citep{wu2025spatialmemory}, GEN3C~\citep{ren2025gen3c}, HY-WorldPlay~\citep{sun2025worldplay}, LingBot-World~\citep{team2026advancing}, Matrix-Game 3.0~\citep{wang2026matrixgame3}, Lyra 2.0~\citep{shen2026lyra2}, and SANA-WM~\citep{zhu2026sana}. For a fair comparison, SPMem~\citep{wu2025spatialmemory}, GEN3C~\citep{ren2025gen3c}, and Lyra 2.0~\citep{shen2026lyra2}, which rely on estimated geometry, use the same Depth Anything 3~\citep{lin2025depth} as our model. All models generate the 60-second video autoregressively with their full denoising schedules, except SANA-WM~\citep{zhu2026sana}, which generates the whole video bidirectionally.

\textbf{Quantitative results.}
We show the quantitative comparisons in Table~\ref{tab:main}. Our model achieves the best results on both benchmarks across nearly all metrics. For camera following and geometric consistency, our model has the lowest translation and reprojection errors on both benchmarks, and reduces the translation error on World Explore from 1.32\,m of the strongest baseline to 1.05\,m. For visual quality, our model has the lowest FID and the highest imaging quality on both benchmarks, and reduces FID from 25.54 to 21.74 on DL3DV-Evaluation and from 42.73 to 35.37 on World Explore. On DL3DV-Evaluation, the generated frames are compared with the real scene along a trajectory that returns to earlier viewpoints, so this gain also reflects the long-term memory of our model. Our model also has the highest text consistency on both benchmarks, a property that the director in Section~\ref{sec:agentic_framework} relies on when prompting the model.

\textbf{Qualitative results.}
Figure~\ref{fig:comparison_video_gen} compares the generated frames on DL3DV-Evaluation and World Explore. Full videos are on our project page.

% Table 2: component ablation on DL3DV-Evaluation, numbers from the 2026-09-15 run (paper_materials/03_experiments/table2_ablation/).
\begin{table}[t]
\caption{Ablation on DL3DV-Evaluation. Metrics are computed over the full 60\,s sequence and over only the last 30, 20 and 10\,s, so the later columns measure long-term consistency. Trans., Rot. and Repr. are translation (m), rotation ($^\circ$) and depth reprojection errors. Occ.: occlusion-aware history selection. Rel.: reliability-based memory maintenance. Best in \textbf{bold}.}
\label{tab:ablation}
\centering
\scriptsize
\setlength{\tabcolsep}{1.3pt}
\begin{adjustbox}{max width=\linewidth}
\newcolumntype{C}{w{c}{19pt}}  % every metric column has the same width
\begin{tabular}{@{}cc *{16}{C}@{}}
\toprule
 & & \multicolumn{4}{c}{Full} & \multicolumn{4}{c}{Last 30\,s} & \multicolumn{4}{c}{Last 20\,s} & \multicolumn{4}{c}{Last 10\,s} \\
\cmidrule(lr){3-6} \cmidrule(lr){7-10} \cmidrule(lr){11-14} \cmidrule(lr){15-18}
Occ. & Rel. & FID$\downarrow$ & Trans.$\downarrow$ & Rot.$\downarrow$ & Repr.$\downarrow$ & FID$\downarrow$ & Trans.$\downarrow$ & Rot.$\downarrow$ & Repr.$\downarrow$ & FID$\downarrow$ & Trans.$\downarrow$ & Rot.$\downarrow$ & Repr.$\downarrow$ & FID$\downarrow$ & Trans.$\downarrow$ & Rot.$\downarrow$ & Repr.$\downarrow$ \\
\midrule
           &            & 22.32 & 0.35 & 1.18 & 0.023 & 34.94 & 0.36 & 1.45 & 0.031 & 41.91 & 0.40 & 1.63 & 0.040 & 57.74 & 0.46 & 2.18 & 0.049 \\
\checkmark &            & 22.66 & 0.39 & 1.06 & 0.022 & 35.41 & 0.42 & 1.22 & 0.028 & 41.70 & 0.47 & 1.23 & 0.033 & 56.94 & 0.56 & 1.40 & 0.040 \\
\checkmark & \checkmark & \textbf{21.74} & \textbf{0.29} & \textbf{0.99} & \textbf{0.020} & \textbf{32.21} & \textbf{0.30} & \textbf{1.11} & \textbf{0.024} & \textbf{37.42} & \textbf{0.33} & \textbf{1.18} & \textbf{0.027} & \textbf{51.84} & \textbf{0.37} & \textbf{1.38} & \textbf{0.031} \\
\bottomrule
\end{tabular}
\end{adjustbox}
\end{table}

\textbf{Ablation.}
Table~\ref{tab:ablation} reports FID and the camera and geometry errors over the full 60 seconds and over the last 30, 20, and 10 seconds, since both components address occlusion and error accumulation over long rollouts, whose effect concentrates in the later part of the sequence. The full model is the best on all four metrics in every window. Removing both components raises FID from 21.74 to 22.32 over the full sequence but from 51.84 to 57.74 over the last 10 seconds. The camera and geometry errors show the same trend. Over the last 10 seconds, removing both components raises the translation, rotation, and reprojection errors from 0.37\,m, 1.38$^\circ$, and 0.031 to 0.46\,m, 2.18$^\circ$, and 0.049. Reliability-based memory maintenance accounts for most of the FID gap (56.94 with occlusion-aware history selection alone), and the occlusion-aware history selection on its own lowers FID only over the last 20 and 10 seconds, since occlusion reasoning depends on the estimated geometry of the history, including its unreliable parts when reliability-based memory maintenance is off. With occlusion-aware history selection alone, the rotation and reprojection errors already drop in every window, but the translation error rises, and adding reliability-based memory maintenance brings it from 0.56 down to 0.37\,m over the last 10 seconds. Figure~\ref{fig:comparison_ablation} shows the ablation variants qualitatively.

\subsection{World Explore Benchmark}
\label{subsec:world_explore_benchmark}

World Explore evaluates world exploration, where the camera keeps moving into parts of the scene that the input image does not show. Its 40 scenes are all videos of a hold-out split of the outdoor part of the internal collection (Section~\ref{subsec:training_data}), and this split was drawn at random from the videos of that part. They are never used in training. We chose the outdoor hold-out over the indoor one because its camera trajectories move away from the start instead of circling the scene, and did not select individual scenes by their camera motion. The videos are mostly handheld or gimbal walks along paths, streets, lakeshores, playgrounds and building exteriors, several of them in snow (Figure~\ref{fig:world_explore_benchmark}, left).

The trajectories of the two benchmarks have similar lengths, with a median of 28.9\,m on DL3DV-Evaluation and 28.4\,m on World Explore, but they cover the scene differently. On DL3DV-Evaluation, the camera stays within a median of 10.0\,m of its start and returns to earlier viewpoints. For World Explore, the camera moves up to a median of 18.1\,m away from its start. The right panel of the figure shows the distance from the start along the trajectories.

\subsection{Agentic World Generation}

We now evaluate the full \name, with \name-Agent directing the world model.

\textbf{Demonstration.}
Figure~\ref{fig:teaser} shows three worlds built by \name. In the 80-second temple video, the camera moves along the main hall, into it, back out and into the right-hand passage, as the instruction asks. Inside the hall, which the input image shows only from outside, \name generates the gilded statue named in the instruction, and the courtyard reappears with the same lanterns and trees. Full videos are on our project page.

\begin{figure}[t]
\begin{center}
\includegraphics[width=0.95\linewidth]{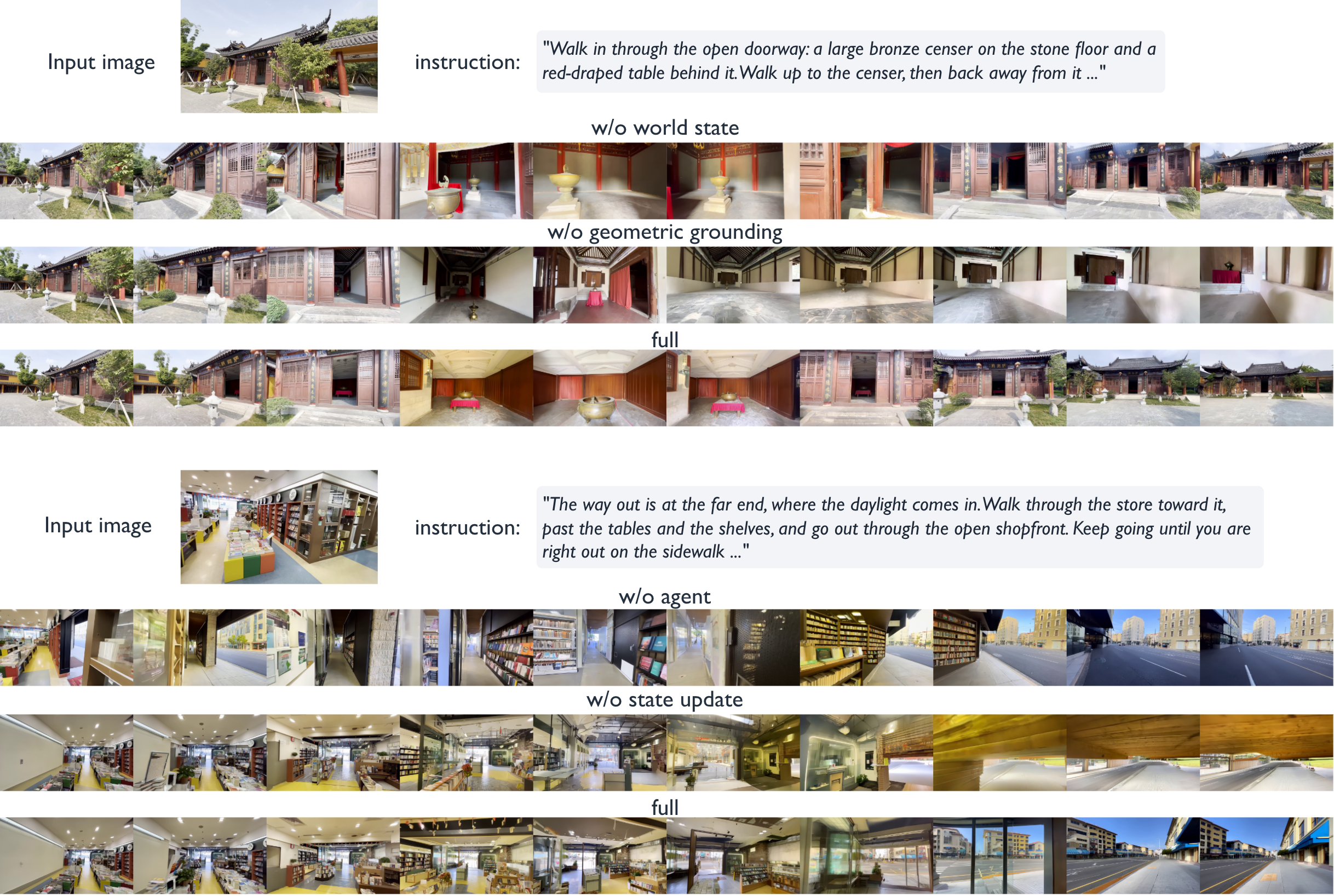}
\end{center}
\caption{Qualitative ablation of \name-Agent on two examples, each shown with its input image and abbreviated instruction. Each row shows frames sampled in order along the video of one variant.}
\label{fig:agent_ablation}
\end{figure}

\textbf{Ablation of \name-Agent.}
We ablate different modules of \name-Agent: the agent as a whole, the world state, the state update, and the geometric grounding.
\emph{W/o agent} plans the whole trajectory at once from the input image and prompts every chunk with the instruction.
\emph{W/o world state} plans each chunk from what has been generated so far, without keeping a world state.
\emph{W/o state update} plans once from the initial world state and generates the whole video without review or state update.
\emph{W/o geometric grounding} uses neither the point cloud nor the geometric tools.
All variants share the input image, instruction, metric coordinate frame and world model.
In the temple of Figure~\ref{fig:agent_ablation}, where the camera should walk up to a bronze censer and back out, the objects in the hall change from one chunk to the next without the world state, and the camera returns to a facade with two doorways instead of one. W/o geometric grounding generates a gallery deeper than the building and never leaves it. In the bookstore, where the camera should walk out to the sidewalk, w/o agent ends in the road with bookshelves standing on the sidewalk, and w/o state update stays under a wooden structure at the shopfront. The full system carries out both instructions. The world state thus keeps the designed world consistent across chunks, geometric grounding matches the plan to the generated geometry, and the state update lets the agent correct its plan after each chunk.

\begin{figure}[t]
\begin{center}
\includegraphics[width=0.95\linewidth]{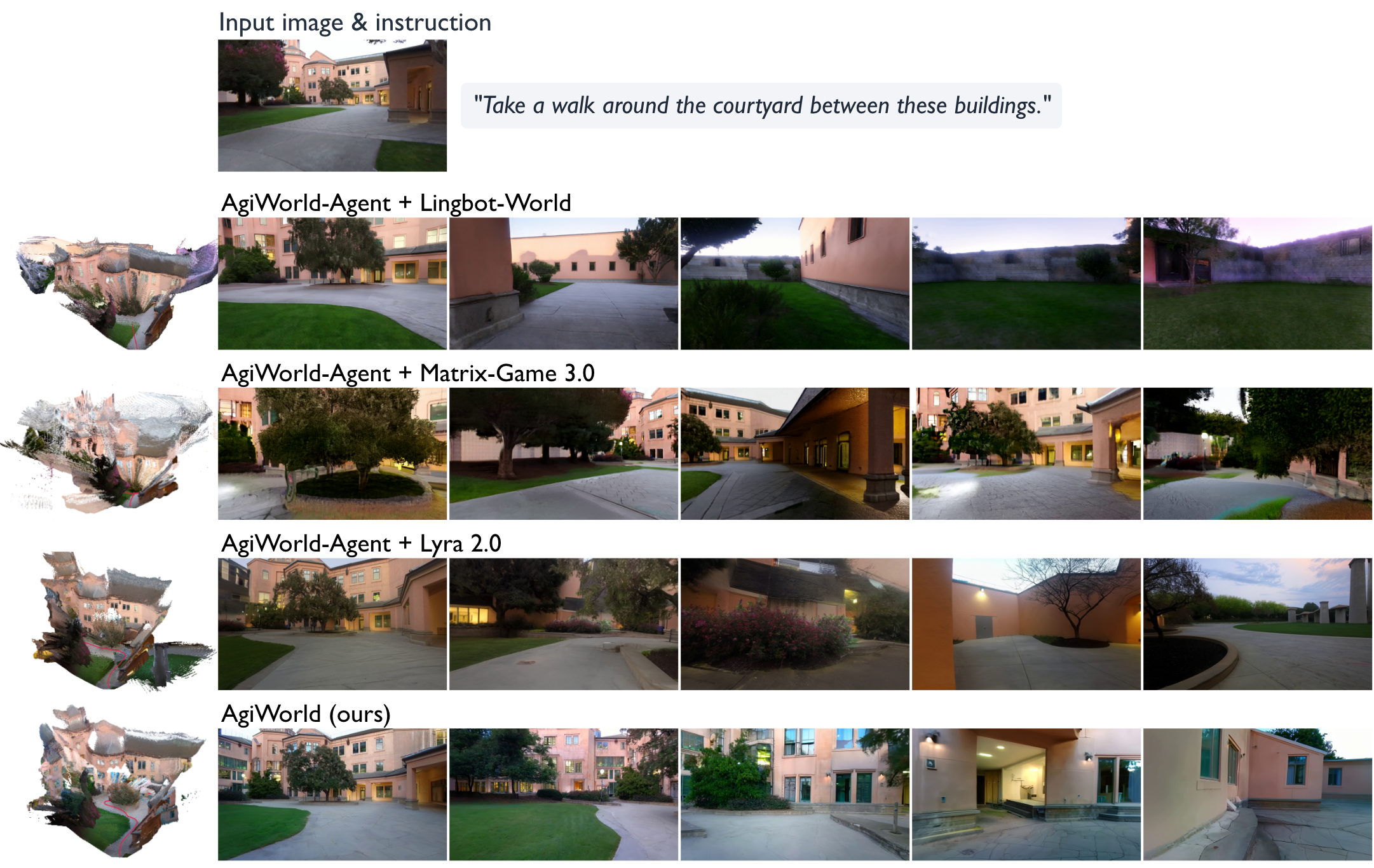}
\end{center}
\caption{Replacing the world model of \name. All rows use the same \name-Agent, input image, instruction and initial world state, and differ only in the world model: LingBot-World, Matrix-Game 3.0, Lyra 2.0 and \name-WM from top to bottom.}
\label{fig:agent_wm_ablation}
\end{figure}

\textbf{Replacing the world model.}
In Figure~\ref{fig:agent_wm_ablation}, we keep \name-Agent with the same inputs and initial world state, and replace only \name-WM with LingBot-World~\citep{team2026advancing}, Matrix-Game 3.0~\citep{wang2026matrixgame3} or Lyra 2.0~\citep{shen2026lyra2}. The agent adapts its plan to each model's output. With \name-WM, all five frames stay in the courtyard of the input image and keep its lighting. The other models reach a purple-tinted stone wall (LingBot-World), blurred vegetation (Matrix-Game 3.0) or an open landscape (Lyra 2.0). The agent can thus direct other camera-controlled world models, while visual persistence comes from \name-WM, which is co-designed with it.

%% file: section/2_related_works.tex
\section{Related Work}

\subsection{Camera-Controlled Video Generation}

Camera-controlled video generation conditions a pretrained video diffusion model on a camera trajectory, and existing methods differ mainly in how the trajectory enters the network. Early methods inject the pose parameters directly, either as flattened extrinsics~\citep{wang2024motionctrl} or as per-pixel Pl\"ucker rays~\citep{sitzmann2021lfn,chen2023rayconditioning} passed through an adapter~\citep{he2024cameractrl,xu2024camco,bahmani2025ac3d,bahmani2025vd3d,he2025cameractrl2}. Recent interactive world models~\citep{he2025matrixgame2,li2025gamecraft,mao2026yume15,hong2025relic} expose camera control as keyboard actions that are mapped into camera space. Geometry-guided methods instead condition on a rendering of the estimated geometry at the target view, a point cloud lifted from the input frames~\citep{yu2024viewcrafter,ren2025gen3c,huang2025voyager,zhao2025spatia} or a depth-warped frame~\citep{li2025magicworld}, and leave the model to fill in the missing regions. Their precision depends on the estimated geometry, and warping introduces disocclusion holes and errors at depth boundaries~\citep{seo2024genwarp} that the model tends to reproduce. A third line modifies the positional encoding of the transformer itself. GTA~\citep{miyato2024gta} and PRoPE~\citep{li2025prope} encode the relative camera transform between query and key tokens as the positional encoding, whereas PE-Field~\citep{bai2025pefield,bai2026pefield4d} encodes the projected position and depth of reference tokens in the target view. Relative camera encodings~\citep{li2025prope} require no fixed reference frame and generalize to unseen sequence lengths.

\subsection{Long Video Generation}

Persistence has been a long-standing difficulty in long video generation. Autoregressive video diffusion models~\citep{chen2024diffusionforcing,huang2025selfforcing} attend only to a bounded window of recent frames, so a region drifts in content, appearance, and geometry once it leaves the window. Later methods extend the window, either by compressing the history into the context~\citep{zhang2025framepack,hong2025relic,wu2026infiniteworld} or by adaptively selecting a sparse subset of the history as context~\citep{yi2026worldkv,cai2025moc,xu2026wonder}. They keep more of the past, but the history is still organized by time and selected by recency or coarse camera correspondence, so a revisited region is not guaranteed to match what was generated before. Another line of methods organizes the history spatially. Retrieval approaches keep past frames and select those that overlap the current view~\citep{yu2025contextasmemory,xiao2025worldmem,sun2025worldplay,wang2026matrixgame3,li2025vmem}, and explicit approaches accumulate a 3D representation and render it into the target view~\citep{wu2025spatialmemory,huang2025voyager}. Matrix-Game~3.0~\citep{wang2026matrixgame3}, for example, retrieves memory frames by camera pose and field-of-view overlap and processes them jointly with the recent history in the same self-attention; Voyager~\citep{huang2025voyager} jointly generates aligned RGB and depth video conditioned on existing world observations, and maintains an efficient world cache with point culling for iterative scene extension. Occlusion and noise in the retrieved or reconstructed content remain difficult to handle, and geometric errors accumulate over long rollouts.

\subsection{Agents for World Generation}

LLM agents can pursue a goal by planning, calling tools, and revising the plan based on the outcomes of their actions~\citep{yao2023react,park2023generativeagents}. These abilities have been brought to content generation. In 3D scene generation, agents can be used to plan the scene as object layouts that are then populated with assets~\citep{feng2023layoutgpt,yang2024holodeck,hu2024scenecraft,ling2025scenethesis}. In VideoDirectorGPT~\citep{lin2024videodirectorgpt}, an agent expands a prompt into multi-scene plans with scene descriptions and entity layouts that a video generator renders scene by scene. MovieAgent~\citep{wu2025movieagent} assigns director and screenwriter roles to agents to produce multi-shot films with consistent characters. Moreover, agents can take control of a visual generation model to automate world generation, deciding what should appear next and steering the generator through text conditions~\citep{yu2024wonderjourney,feng2025wonderverse,gao2026lingbotworld2}. WonderJourney~\citep{yu2024wonderjourney} leverages an LLM to generate textual descriptions of the scenes along a journey, an inpainting model with depth estimation to produce each scene as a point cloud, and a VLM to verify the generated scenes. LingBot-World~2.0~\citep{gao2026lingbotworld2} pairs its autoregressive video generator with an agent that provides the input actions and next-event control signals in place of the user.

%% file: section/8_conclusion.tex
\section{Conclusion}

In this work, we proposed \name, a novel agentic video world model that builds a persistent world from a single image and a free-form instruction. We introduced \name-Agent, which designs the world beyond the current view and the journey through it in a high-level world state, directs the world model with metric camera trajectories and structured prompts, and revises its design in a closed loop as the world is generated. We designed \name-WM, a camera-conditioned autoregressive video world model whose spatial memory stays consistent over long rollouts through novel occlusion-aware history selection and reliability-based memory maintenance. We co-designed the two around this shared spatial memory and the interface between them. \name-WM achieves state-of-the-art overall performance on world revisiting and exploration benchmarks, and \name builds persistent worlds that follow free-form instructions over minute-long journeys. We believe that \name opens a new direction for video world models, in which an agent plans the world and directs the world model to generate it, instead of a user steering the camera chunk by chunk.

%% file: section/appendix.tex
\beginappendix

\input{section/appendix/agent_framework}

%% file: section/appendix/agent_framework.tex
\section{\name-Agent Details}
\label{appendix:agent_framework}

This section gives the implementation details of \name-Agent (Section~\ref{sec:agentic_framework}), in the order of the main text.

\subsection{Overview}
\label{appendix:agent_overview}

Section~\ref{subsec:closed_loop} defines the two functions of \name-Agent. The director $\mathcal{D}$ turns the world state into the trajectory and the prompt of each chunk, and the state update $\mathcal{U}$ updates the world state from the memory of \name-WM and designs the world and the journey ahead. Both functions are carried out by one multimodal language model, GPT-6-Astra in all our experiments, together with the geometric tools. The agent does not depend on this model, and other multimodal language models, such as Claude Opus, Claude Fable, and GPT-5.6, can take its place. The language model is called in separate steps, each with its own guidelines and output schema (Table~\ref{tab:agent_steps}). The tools prepare the inputs of each step and check its output.

\textbf{Director.} The director $\mathcal{D}$ consists of four steps. Three of them plan the route. The trajectory step writes waypoints for the next 60\,s, and the tools solve them into a camera path. The fix step corrects the waypoints when this path fails a check, and the prompt step writes the prompts of the chunks along the path. These three steps run at initialization and again whenever the route has to change, which we call a replan. The fourth step, the review, checks each generated chunk against the plan and reads the camera measurements that the tools compute on it. A chunk whose camera fails these measurements, or that shows something else in place of an element named in the instruction, is generated again. The small edits of the layout and the stages that the review returns are passed to the state update.

\textbf{State update.} The state update $\mathcal{U}$ has two parts. Reconstruction writes the confirmed layout from the input image and the generated geometry, and design writes the journey and the imagined layout. Each part has one step at initialization and one step that can run after each chunk. At initialization, the state update $S_1=\mathcal{U}(\varnothing,\mathcal{M}_1,g_0)$ runs the initial reconstruction of the scene visible in $I_0$, which reads the ground, writes the layout, and verifies it. The initial design of the journey and of the world beyond the image follows. After each chunk, the state update $S_{k+1}=\mathcal{U}(S_k,\mathcal{M}_{k+1},g_0)$ applies the edits of the review, revises the confirmed layout when the chunk contradicts it, which we call a revision, and rewrites the stages and the imagined layout when they have to change, which we call a refresh.

\textbf{Queries.} Each query contains the guidelines of the step, the images and data that the geometric tools prepared for it, and a JSON output schema. The queries of the initial design, the refresh, and the trajectory step also contain a shared specification of \name-WM. The full guidelines, schemas, and specification will be released with the code.

\textbf{Validation.} The tools check each output against its schema and against the checks of its step. An output that fails is returned with a list of its errors, and the model answers again. When a revision, a refresh, or a replan still fails, the loop keeps the previous state and continues. A path that still violates a bound after the fix step is the exception and is accepted.

\textbf{Order of calls.} Algorithm~\ref{alg:agent_loop} gives the order of all steps. The review and the state update run after every chunk. Regeneration, revision, refresh, and replan run only when their triggers fire.

\begin{figure}[t]
\begin{center}
\includegraphics[width=0.95\linewidth]{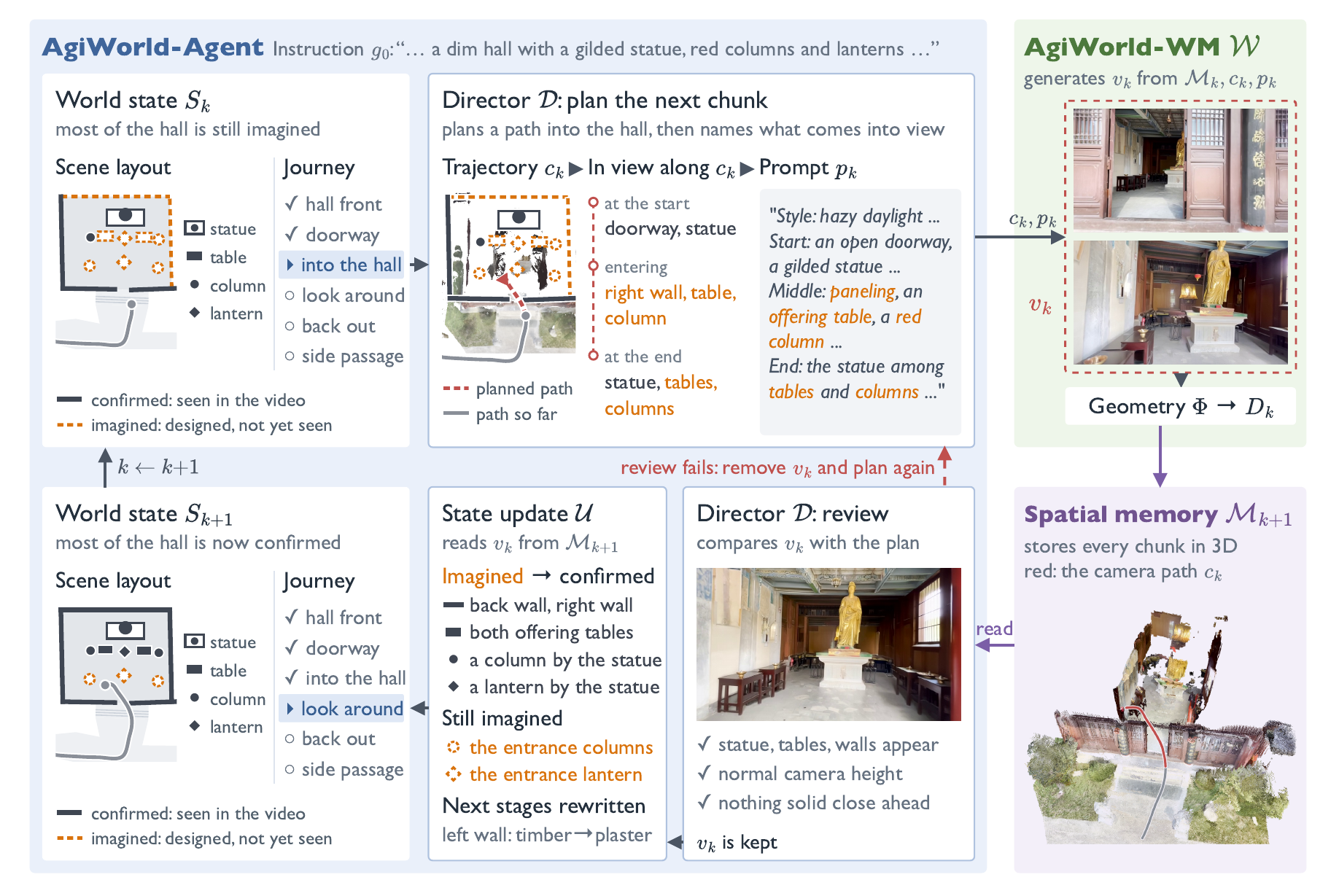}
\end{center}
\caption{All decisions and state changes of \name-Agent for one chunk, from the temple example in Figure~\ref{fig:teaser}: the director plans the chunk, \name-WM generates it, the director reviews it, and the state update turns $S_k$ into $S_{k+1}$.}
\label{fig:agentic_detailed_framework}
\end{figure}

\begin{table}[ht]
\centering
\caption{The steps of \name-Agent, grouped by the function they carry out. Each step has its own guidelines and output schema.}
\label{tab:agent_steps}
\small
\begin{tabular}{@{}>{\raggedright\arraybackslash}p{0.2\linewidth}>{\raggedright\arraybackslash}p{0.42\linewidth}>{\raggedright\arraybackslash}p{0.31\linewidth}@{}}
\toprule
Step & Called & Writes \\
\midrule
\multicolumn{3}{@{}l}{\textit{Director $\mathcal{D}$}} \\
Trajectory & At initialization, and whenever the route has to change (replan) & Waypoints of the next 60\,s \\
Fix & After Trajectory, when the solved path fails a check & Corrected waypoints \\
Prompt & After Trajectory and Fix & Start, middle, and end of the prompt of each chunk \\
Review & After every chunk & Verdicts, small layout edits, stage updates, decision for the next chunk \\
\midrule
\multicolumn{3}{@{}l}{\textit{State update $\mathcal{U}$: reconstruction}} \\
Initial reconstruction & At initialization: ground reading, layout, verification & Standing surface and other marked surfaces, confirmed layout \\
Revision & Major mismatch, contradicted target, or ground offset & Confirmed layout, contradicted imagined elements \\
\midrule
\multicolumn{3}{@{}l}{\textit{State update $\mathcal{U}$: design}} \\
Initial design & At initialization & Style, stages, imagined layout \\
Refresh & Stage status changed, less than 30\,s of path left, layout revised, or elements retired & Stages, edits of the imagined layout, finished flag \\
\bottomrule
\end{tabular}
\end{table}

\begin{algorithm}[t]
\caption{The generation loop of \name. LM marks calls to the language model, WM marks calls to \name-WM and its geometry estimator $\Phi$, and tools marks deterministic computation.}
\label{alg:agent_loop}
\small
\begin{algorithmic}[1]
\Require image $I_0$, instruction $g_0$
\State $\mathcal{M}_1 \gets \{(I_0,\mathbf{T}_0,\Phi(I_0))\}$ \Comment{WM}
\State Initial reconstruction: read the ground, write and verify the confirmed layout \Comment{LM, tools}
\State Initial design: write the style, the stages, and the imagined layout \Comment{LM, tools}
\State $S_1 \gets \mathcal{U}(\varnothing,\mathcal{M}_1,g_0)$, the state written in lines 2--3
\State Director: trajectory, fix, and prompt steps for the first 60\,s \Comment{LM, tools}
\State $k \gets 1$
\Repeat
  \State $(c_k,p_k) \gets \mathcal{D}(S_k,g_0)$, taken from the planned path and prompts
  \State $v_k \gets \mathcal{W}(\mathcal{M}_k,c_k,p_k)$ \Comment{WM}
  \State $D_k \gets \Phi(v_k,c_k)$, \quad $\mathcal{M}_{k+1} \gets \mathcal{M}_k \cup \{(v_k,c_k,D_k)\}$ \Comment{WM}
  \State Measure the camera on the generated geometry \Comment{tools}
  \State Director: review $v_k$ \Comment{LM}
  \If{$k>1$ and a regeneration check fires} \Comment{tools}
    \State Restore $\mathcal{M}_k$ and $S_k$ \Comment{WM, tools}
    \State Director: replan from the end of chunk $k-1$, given the first review \Comment{LM, tools}
    \State Repeat lines 8--12 once, with a new seed
  \EndIf
  \State If the review rewrote the next prompt, replace $p_{k+1}$
  \State $S_{k+1} \gets \mathcal{U}(S_k,\mathcal{M}_{k+1},g_0)$: apply the edits of the review, then revise and refresh if triggered \Comment{LM, tools}
  \State If the route has to change, the director replans with the trajectory, fix, and prompt steps \Comment{LM, tools}
  \State $k \gets k+1$
\Until{a refresh marks the journey as finished}
\end{algorithmic}
\end{algorithm}

\subsection{World State}
\label{appendix:agent_state}

Table~\ref{tab:agent_state} lists the fields of the world state.

\textbf{Coordinate frame.} All positions are metric and lie in one frame. It is the frame of $I_0$, leveled to gravity by the ground reading at initialization, and $\mathbf{T}_0$ denotes the pose of $I_0$ in this frame. The first camera lies at $x=z=0$. The $x$ axis points to the right of the first camera, the $z$ axis points forward, and both are horizontal. The height $h$ is measured upward from the standing surface under the first camera. The camera poses $c_k$ that \name-WM receives and the geometry of $\mathcal{M}_k$ are expressed in the same frame.

\textbf{Layout.} The layout is a set of named elements. The initial reconstruction and each revision write the confirmed elements, and the review may add, confirm, revise, or drop elements in small edits. The initial design and each refresh write the imagined elements. An imagined element becomes confirmed when the review marks it as generated as designed, or when a revision includes it with its observed geometry. An imagined element is retired when a revision reports that the video contradicts it, or when the confirmed layout grows over its place. Retired elements are kept as a record so that they are not proposed again.

\textbf{Journey.} The initial design writes the journey, and each refresh rewrites the stages. The review marks a stage as done once a chunk has shown its content. A stage states what the camera should show and contains no coordinates or speeds, because the route and the pace are decided by the director.

\begin{table}[ht]
\centering
\caption{Fields of the world state.}
\label{tab:agent_state}
\small
\begin{tabular}{@{}>{\raggedright\arraybackslash}p{0.2\linewidth}>{\raggedright\arraybackslash}p{0.76\linewidth}@{}}
\toprule
Field & Meaning \\
\midrule
\multicolumn{2}{@{}l}{\textit{Layout element}} \\
Name & What the element is, with a fixed feature of the place when several are alike, e.g., ``offering table on the passage side''. Words that depend on the camera, such as \emph{left} or \emph{ahead}, are not allowed. \\
Kind & Surface, terrain, structure, object, opening, or zone. Surfaces and terrain form the ground, and structures and objects are solids. \\
Shape & Polygon, polyline, path, box, circle, or point, with metric coordinates in the $(x,z)$ plane and an optional elevation per vertex for ramps and stairs. \\
Elevation, height & Height of the base above $h=0$, and vertical extent. \\
Appearance & One line that describes what the element looks like. \\
Label & Confirmed (generated and observed) or imagined (designed by the agent but not yet reached by the camera). \\
Flags & Walkable, named in the instruction, read from $I_0$. \\
\midrule
\multicolumn{2}{@{}l}{\textit{Retired element}} \\
Record & Name, kind, position, and when and why it was retired. \\
\midrule
\multicolumn{2}{@{}l}{\textit{Journey}} \\
Summary & A few sentences on what the place is and how the journey proceeds. \\
Style & A medium, such as photorealistic real-world footage, and a look that names the light, the shadows, and the palette. Fixed for the whole journey. \\
Stages & Ordered list. Each stage has a goal, what is on screen and from where, a soft time window, the elements it uses, and a status (pending, active, done, or skipped). \\
Finished & Set by a refresh when the journey is complete. It ends the loop. \\
\bottomrule
\end{tabular}
\end{table}

\subsection{Geometric Grounding}
\label{appendix:agent_grounding}

\begin{figure}[t]
\begin{center}
\includegraphics[width=\linewidth]{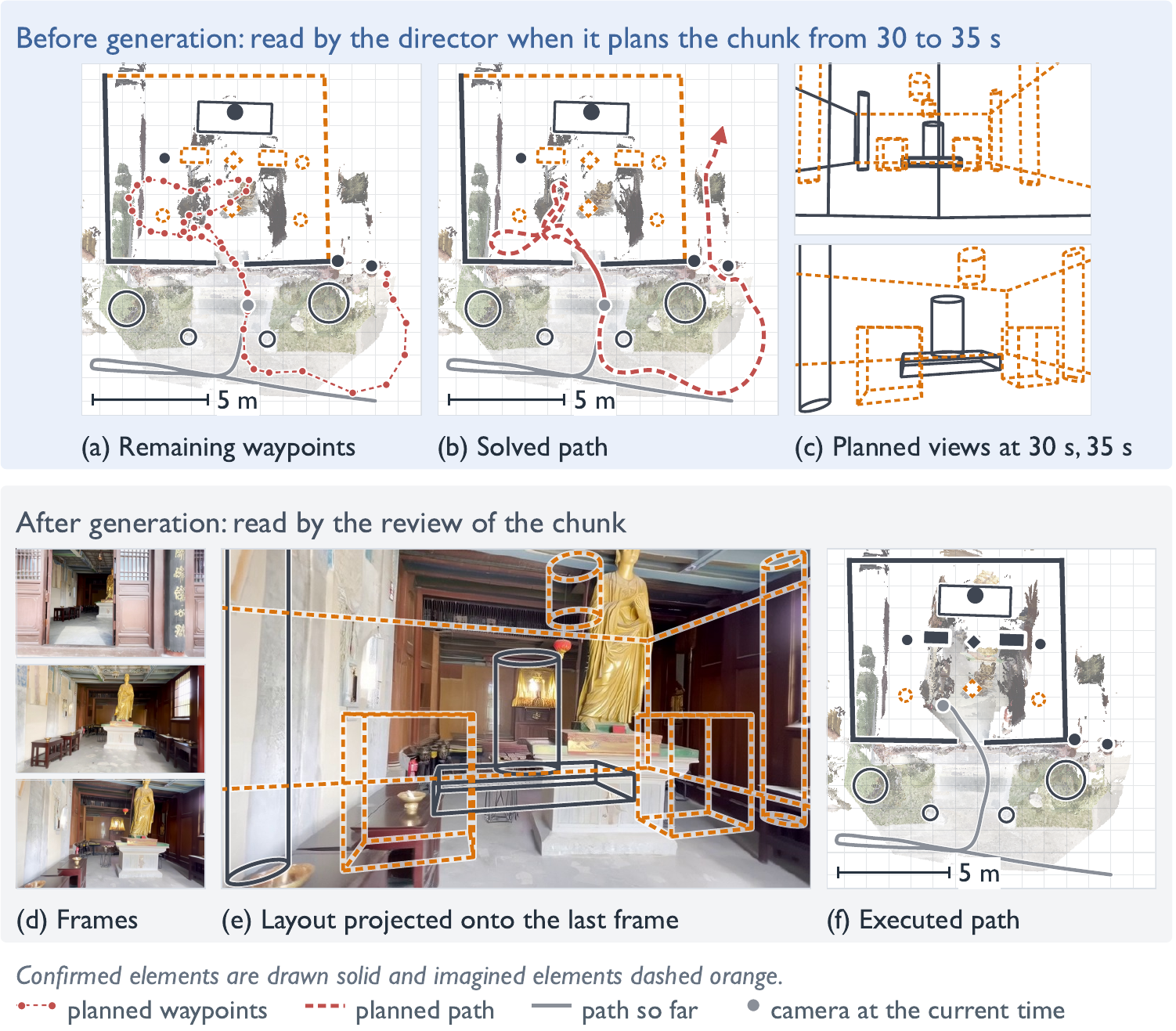}
\end{center}
\caption{Renders prepared by the geometric tools in one iteration, redrawn from the intermediate results of the temple example in Figure~\ref{fig:teaser}, with the point cloud in its image colors and all confirmed elements in one color. The iteration generates the chunk from 30 to 35\,s, in which the camera walks into the temple hall, and its path comes from a replan after the previous chunk. Top, before the chunk is generated: (a) the floor plan of the layout with the remaining planned waypoints, read by the trajectory step, (b) the solved path, read by the fix step, and (c) the planned views at 30\,s and 35\,s, read by the prompt step. Bottom, after the chunk is generated: (d) frames of the chunk, (e) the layout projected onto its last frame, and (f) the floor plan of the accumulated point cloud with the executed path, all read by the review. Confirmed elements are drawn solid and imagined elements dashed.}
\label{fig:agent_observations}
\end{figure}

Table~\ref{tab:agent_tools} lists what the geometric tools compute and render, and which steps read each output. Figure~\ref{fig:agent_observations} shows the renders of one iteration.

\textbf{Sources of geometry.} At initialization, the geometry estimator $\Phi$ estimates the depth of $I_0$. During the rollout, the tools read the depth $D_k=\Phi(v_k,c_k)$ stored in $\mathcal{M}_{k+1}$. They accumulate these depths and the depth of $I_0$ into one point cloud. They keep the points with a confidence of at least 0.35 within 25\,m of their camera, and remove the sky.

\textbf{Ground reading.} A plane fitted to the whole image fails to find the ground when little of it is visible. The initial reconstruction therefore reads the ground first, before it writes the layout. To read the ground, the language model draws boxes on $I_0$ around the standing surface under the camera, other horizontal surfaces, and vertical faces such as walls or columns. The tools fit planes to the depth inside these boxes. The up direction comes from the horizontal surfaces, or from the vertical faces when no horizontal surface gives a usable plane. The floor $h=0$ comes from the standing surface under the camera. When no marked surface gives a usable plane, or when the standing surface slopes by more than 15$^\circ$ or the ground 1 to 6\,m ahead lies less than 0.2\,m below the eye, the tools fall back to a plane fitted to the whole image with RANSAC.

\textbf{Visibility.} To find what each chunk will show, the tools project every layout element along the solved path. Each face of an element is rasterized with its depth, so structures and objects hide what lies behind them. A declared opening cuts a hole in the wall that contains it, so an element behind a doorway is visible through the doorway.

\textbf{Drawing conventions.} All renders share the frame of the layout. Confirmed elements have solid outlines colored by kind, and imagined elements have dashed orange outlines. Floor plans, the top-down maps of the main text, show the point cloud from above, colored by height, on a metric grid, with the first camera at the origin, the executed path in gray, and the planned path in red. Projections into an image draw each element as a wireframe from its base to its top.

\begin{table}[ht]
\centering
\caption{Outputs of the geometric tools and the steps that read them.}
\label{tab:agent_tools}
\small
\begin{tabular}{@{}>{\raggedright\arraybackslash}p{0.2\linewidth}>{\raggedright\arraybackslash}p{0.47\linewidth}>{\raggedright\arraybackslash}p{0.26\linewidth}@{}}
\toprule
Tool output & Content & Read by \\
\midrule
Depth map of $I_0$ & The depth of $I_0$ as a color map with a metric scale. & Initial reconstruction \\
Point cloud and samples of $I_0$ & The leveled point cloud of $I_0$ from above, and $I_0$ with a grid of points labeled with their $(x,z,h)$. & Initial reconstruction \\
Layout check & For each element, whether its position and height agree with the depth and the point cloud of $I_0$. & Initial reconstruction \\
Layout projection & The layout drawn as wireframes into $I_0$ or into the last frame of a chunk. & Initial reconstruction, Revision, Review \\
Floor plan & The layout over the accumulated point cloud, with the executed path and the planned path. & All steps except Prompt \\
Trajectory check & Collisions, eye-height and support violations, and violations of the motion bounds on the solved path. & Fix \\
Visibility timeline & The elements in view along each planned chunk. & Prompt \\
View previews & The planned view at each chunk boundary, with the element wireframes. & Prompt \\
Chunk frames & Five evenly spaced frames of the latest chunk. & Revision, Refresh, Trajectory (replan), Review \\
Camera measurements & The camera measured on the generated geometry. & Review, Revision, Refresh, Trajectory (replan) \\
\bottomrule
\end{tabular}
\end{table}

\subsection{Director: Trajectory}
\label{appendix:agent_trajectory}

\textbf{Waypoints.} The trajectory step writes sparse waypoints on the floor plan. Each waypoint has a time, a position $(x,z)$, a camera height $h$, a look-at target, the stage it serves, and a one-line purpose. The look-at target is a layout element, a point, or a heading with an optional pitch.

\textbf{From waypoints to $c_k$.} The tools turn each waypoint into a look-at pose with zero roll. They interpolate the poses, starting from the current pose, into a $C^1$ curve that starts with the velocity of the executed path, so a new path continues the executed one without a jump. The curve is sampled at 16\,fps, and we call it the path. Each chunk takes 80 consecutive poses of the path, which cover 5\,s and form its trajectory $c_k=\{\mathbf{T}_{k,f}\}_{f=1}^{n}$ with $n=80$. Each request to \name-WM contains these metric poses, the camera intrinsics that $\Phi$ estimated for $I_0$, which stay fixed for the whole rollout, and the prompt $p_k$.

\textbf{Planning window.} The director designs the path for the next 60\,s. When less than 30\,s of the path remain, a refresh extends the stages, and the director replans the path for the next 60\,s. The director replans from the current pose whenever the route has to change.

\textbf{World-model specification.} The initial design, the refresh, and the trajectory step share one specification of \name-WM in their guidelines. It states what the model does well and what it cannot do, together with the camera and prompt practices that follow from each (Table~\ref{tab:agent_wm_spec}). We wrote it from the behavior of the model observed in generated videos. The specification also asks for a walking pace of about 0.6 to 0.9\,m/s and for an eye height above the walked surface equal to that of $I_0$. The view tilts up or down only in a chunk that shows something above or below that the instruction names.

\begin{table}[ht]
\centering
\caption{The specification of \name-WM shared by the initial design, the refresh, and the trajectory step. Each practice follows from the behavior on its left.}
\label{tab:agent_wm_spec}
\small
\begin{tabular}{@{}>{\raggedright\arraybackslash}p{0.44\linewidth}>{\raggedright\arraybackslash}p{0.52\linewidth}@{}}
\toprule
Behavior of \name-WM & Camera and prompt practice \\
\midrule
It extends what is already in the middle of the frame while the camera moves smoothly and most of what the previous seconds showed stays in view. & The camera turns gently while it walks, so that most of what the last five seconds showed stays in view. It walks in the direction it looks, and walks backward only for a few seconds. \\
\midrule
A space that it has never shown is hard to generate at the moment the camera enters it. Walking through a doorway or around a corner onto unseen ground can produce a jump. & A new space is kept in the middle of the frame from the first chunk in which it is visible, and the camera walks toward it without stopping. The prompt of each of these chunks states what is there. The chunk that crosses an opening ends well beyond it. \\
\midrule
Its memory holds only what it has generated. Parts of a room that were never in view have nothing to stay consistent with. & In a room, the first chunk turns the view across the room, about a quarter turn while walking slowly, so that its walls are in the memory before the camera leaves them. To look back, the camera turns around while walking. \\
\midrule
It keeps a large space most consistent when the space is shown whole from a distance. Narrow views that each cover a small patch of a room give it little to keep consistent. & Large scenes are shown from far. An element that the instruction names is held in the middle of the frame, from far enough back that the space around it is also in view. \\
\midrule
A wall or a shelf that fills the frame from close range gives it nothing to continue. & The camera looks where the space ahead opens, down a lane or across a room, rather than at the wall it passes. It moves sideways only while the view stays open. \\
\midrule
It generates what the prompt names where the frame gives it room, and what it generates can differ from the layout. & The prompt names only what the frames of the chunk will certainly show. The next path is designed on the generated geometry. \\
\bottomrule
\end{tabular}
\end{table}

\textbf{Checks.} Each time the trajectory step writes waypoints, the tools solve them and check every chunk of the path against the bounds of Table~\ref{tab:agent_bounds}. These bounds are looser than the practices of the specification. The estimated depth often places a surface across a doorway, so the point cloud can close an opening that the camera can pass. The collision check therefore ignores the point cloud inside declared openings. A solved path that violates a bound returns to the director in the fix step. After at most three rounds of correction, the last path is accepted even if a bound is still violated.

\begin{table}[ht]
\centering
\caption{The main bounds on the solved path. A violation is returned to the director.}
\label{tab:agent_bounds}
\small
\begin{tabular}{@{}>{\raggedright\arraybackslash}p{0.2\linewidth}>{\raggedright\arraybackslash}p{0.76\linewidth}@{}}
\toprule
Check & Bound \\
\midrule
Collision & A sphere of radius 0.25\,m around the camera stays clear of the confirmed solids and of the point cloud. \\
Eye height & The eye stays within 0.7 to 1.4 times its starting height above the walked surface of the layout. \\
Support & The camera does not pass over an area that the layout marks as not walkable. \\
Speed & At most 2.5\,m/s, with a change of pace of at most 0.6\,m/s$^2$. \\
Turning & At most 90$^\circ$/s, and at most 25$^\circ$/s on average over a chunk. \\
Chunk boundary & Between consecutive planned chunks, the velocity changes by at most 1\,m/s and the turn rate by at most 45$^\circ$/s. \\
Travel & At least 1.5\,m in every chunk. \\
\bottomrule
\end{tabular}
\end{table}

\subsection{Director: Prompt}
\label{appendix:agent_prompt}

\textbf{Visibility along the path.} The prompt step derives the prompt $p_k$ from the solved path. After the path is accepted, it writes the prompts of all chunks in the planning window at once. For each chunk, the tools give it four lists: the elements visible in the first frame, the elements that enter the view or grow, the elements that leave the view, and the elements visible in the last frame. Each list is ranked by the largest share of the frame that an element covers during the chunk. Each imagined element is also marked with whether it may be named under the rules below. The prompt step also receives a preview of the planned view at each chunk boundary.

\textbf{Format.} The prompt uses the structured format of the training captions (Section~\ref{sec:data_and_training}): a style line followed by the start, the middle, and the end of the chunk. The director writes the three sections, and the tools put the style line of the journey in front of them. The start describes the first frame, and the end describes the last frame. The middle describes what becomes visible, grows, or leaves the view over the five seconds, and what stays in view. The director aims at 27, 73, and 38 words for the three sections, which are the median lengths of the training captions.

\textbf{Naming rules.} \name-WM generates what the prompt names wherever the frame gives it room. An imagined element may therefore be named only in a chunk that sees it through a confirmed opening, or that holds it in the middle third of the frame for at least 2\,s within 10\,m. Every imagined element that the instruction names and that enters the view under this rule must be named in the middle section of the prompt. A confirmed element that covers less than 2\% of the frame is not described unless the instruction names it or an earlier chunk showed it clearly, because a description makes the model generate it larger than it is. \name-WM also generates what the prompt negates. The prompt therefore contains no negations, except ``no longer visible'' and its variants, which the training captions use for an element that leaves the view. The scene is described as static, and camera words are not allowed. An element named in the instruction is described as the instruction describes it, even after a chunk has generated it differently.

\textbf{Validation.} The tools check the word ranges of the three sections (18 to 40, 50 to 115, and 22 to 60 words), negations, camera words, and the naming rules, also for a prompt that the review rewrites.

\subsection{Director: Review and Regeneration}
\label{appendix:agent_review}

\name-WM follows the planned camera poses, but the world that it generates around them can depart from the design. A camera that keeps its planned height above the layout can still sink toward the generated ground, and a path that clears a designed wall can still pass close to the wall that was actually generated. The tools therefore measure the camera on the geometry that \name-WM estimated for the generated frames, and pass the measurements to the review.

\textbf{Measurements.} The tools measure on the point cloud accumulated from $I_0$ and all chunks generated so far. They evaluate eight poses of the chunk and report the median of each quantity over these poses.
\begin{enumerate}
\item The ground plane is fitted to the points within 8\,m of the camera, measured horizontally, that lie at most 0.35\,m above the 25th percentile of their heights.
\item The eye height is the height of the camera above this plane.
\item The ground beside the path is measured on the points 1 to 6\,m ahead that lie within 0.3\,m of the plane. Its width to the left and to the right of the line of travel is given by the 2nd and 98th percentiles of their lateral offsets.
\item The nearest solid is the 5th percentile of the horizontal distance of the points in a cone of $\pm 30^\circ$ around the view direction, between 0.3 and 2.5\,m above the plane.
\item The tilt is the angle between the view direction and the ground plane.
\end{enumerate}

\textbf{Inputs and outputs.} The director reviews the chunk from one composite image that shows $I_0$, five frames of the chunk, the layout projected onto its last frame, and a floor plan of the accumulated point cloud with the executed path and the next planned segment. It also reads the stages, the prompt of the chunk, the elements named in the instruction, the plan of the next chunk, the measurements, and the targets, which are the imagined elements that the next stages use. For each element named in the prompt, the review reports whether it was generated as described, is absent, or is different, meaning that something else stands in its place. For each target, it reports whether the target is established, not yet visible, or contradicted, for example by a wall or a closed door in its place. The review rates the overall content match as ok, minor, or major. A major mismatch means that the layout around the camera is wrong, or that the camera entered an imagined space that looks different from its design. The review also returns small layout edits (confirm, revise, drop, or add), stage updates, and a decision for the next chunk: keep it, rewrite its prompt, or ask for a new route.

\textbf{Regeneration.} A chunk fails the review, and is generated again, when one of the following checks fires. The eye-height checks compare with a reference eye height measured on the first chunk.
\begin{itemize}
\item The eye height is below 0.55 times or above 1.4 times the reference.
\item The nearest solid ahead is closer than 0.7\,m.
\item Less than 0.25\,m of generated ground lies on one side of the path.
\item The view points more than 10$^\circ$ below the ground plane.
\item The review reports an element named in the instruction as different or contradicted.
\end{itemize}
We set the thresholds on chunks of earlier rollouts, so that no chunk accepted on inspection crosses them. The overall content match alone does not trigger a regeneration, and the other disagreements are resolved by the state update.

When a check fires, the state is restored to its value before the review. \name-WM restores its memory to $\mathcal{M}_k$ and discards the frames, latents, and geometry of $v_k$. The director replans from the end of the previous chunk, given the reasons and the first review, and \name-WM generates the chunk again with a new seed. The new chunk is measured and reviewed like any other.

A new sample often clears a fault that occurred by chance, such as a view that sinks toward the ground, but it rarely changes how the model generates a place. We therefore bound regeneration. The first chunk is never regenerated, because it defines the reference eye height. Each chunk is regenerated at most once, a rollout regenerates at most four chunks, and each element named in the instruction triggers at most one regeneration. A fault that persists after a regeneration no longer triggers a regeneration for the rest of the rollout and is only reported.

\subsection{State Update}
\label{appendix:agent_update}

After every chunk, the state update first applies the small layout edits and stage updates of the review. The revision and the refresh then run only when their triggers fire, and the director replans afterwards when the route has to change.

\textbf{Revision.} The confirmed layout is revised when the review reports a major mismatch or a contradicted target. It is also revised when the generated ground and the surface of the layout differ by more than 0.3\,m under the camera for two consecutive chunks. The revision returns the complete confirmed layout. It keeps the elements that still hold, corrects those that the frames contradict, and adds what the frames newly show. Imagined elements that were generated are promoted to confirmed with their observed geometry. Imagined elements that the video contradicted are listed and retired.

\textbf{Refresh.} A refresh runs when a stage changes status, when less than 30\,s of the path remain, when the confirmed layout was revised, or when imagined elements were retired. A refresh returns the stages with their status, new stages where the journey runs short, and edits of the imagined layout (add, revise, or drop). The refreshed stages have to cover the next 60\,s. A stage is skipped only when the video contradicted its space. The imagined layout of the initial design and of every refresh has to pass four consistency checks against the confirmed layout, and a failed check is returned with its errors.
\begin{enumerate}
\item Imagined ground connects to confirmed ground, directly, through an opening, or through other imagined ground.
\item Imagined ground, openings, and zones do not cross a confirmed wall outside a confirmed opening, and do not lie inside a confirmed solid.
\item Imagined objects do not overlap a confirmed solid at the same height.
\item Imagined ground does not cover confirmed ground at the same level.
\end{enumerate}

\textbf{Retirement.} After the edits of the review and after a revision, the confirmed layout can grow over imagined elements. The tools then retire every imagined element that fails the second or the fourth check, and the retirement triggers a refresh.

\textbf{Replan.} The director replans the route from the current pose when the review asks for a new route, when the layout was revised, when a refresh changed the stages that the route follows or the elements they use, when the remaining path collides with the updated layout, or when less than 30\,s of the path remain. A replan runs the trajectory, fix, and prompt steps again for the next 60\,s.

\textbf{Instruction and input image.} The generated video overrides the design of the agent but not the instruction. Elements read from $I_0$ are not revised or dropped to match a generated chunk. A confirmed element named in the instruction is not dropped or changed in kind. An imagined element named in the instruction is not dropped before a fair attempt. A fair attempt is a chunk in which the camera held the element in the middle third of the frame for at least 2\,s within 10\,m, the middle section of the prompt named it, and the review reported it as absent or different. This bound keeps the journey from stopping at an element that \name-WM does not generate. When the video shows something else in place of a named element, the chunk is regenerated, and until a fair attempt has been made, the director keeps approaching the element and describing it as the instruction does.

\textbf{Termination.} The loop ends when a refresh marks the journey as finished, which it does once the instruction has been carried out.